\documentclass[journal]{IEEEtran}
\usepackage{amsmath,amsfonts}
\usepackage{algorithmic}
\usepackage{algorithm}
\usepackage{array}
\usepackage[caption=false,font=normalsize,labelfont=sf,textfont=sf]{subfig}
\usepackage{textcomp}
\usepackage{stfloats}
\usepackage{url}
\usepackage{verbatim}
\usepackage{graphicx}
\usepackage{cite}
\usepackage{xcolor}
\newcommand{\added}[1]{#1} %

\usepackage{bm}

\def\qut#1{\left(#1\right)}

\def\bmy{\bm{y}}
\def\bmyh{\bm{\hat{y}}}
\def\bmx{\bm{x}}
\def\bmxh{\bm{\hat{x}}}
\def\bmz{\bm{z}}

\def\bmn{\bm{n}}

\def\bbE{\mathbb{E}}

\def\qut#1{{\left( #1 \right)}}
\def\quts#1{{\left[ #1 \right]}} %

\usepackage{amsthm}
\usepackage{amsmath}
\usepackage{graphicx}
\usepackage{makecell}
\def\bbR{\mathbb{R}}
\newtheorem{Proposition}{Proposition}
\newtheorem{Assumption}{Assumption}
\newtheorem*{Definition*}{Definition}
\newenvironment{proofsketch} {\begin{proof}[Proof Sketch]} {\end{proof}}

\usepackage{slashbox}
\usepackage{multirow}
\usepackage{booktabs}

\begin{document}

\title{Learning-based Statistical Refinement for Denoising}

\author{Rihuan Ke %
\thanks{The author is with the School of Mathematics, University of Bristol,
BS8 1QU Bristol, U.K. (e-mail: rihuan.ke@bristol.ac.uk). 
} %
}

\maketitle

\begin{abstract}
This work proposes a learning-based statistical refinement method for improving the denoising results of a given denoiser without knowing the precise noise distribution or accessing clean images or calibration data. While there are many existing successful denoising approaches for handling different kinds of noise, they typically require accurate modelling of the images and the noise (implicitly or explicitly), and hence the denoising results can be suboptimal due to different practical factors such as imperfect models, unreliable noise assumptions, or low quality data. In particular, when clean image samples are not available and there is a lack of knowledge of the underlying noise distribution, which is the case in various practical situations, the results may not well align with the noise statistics. The unawareness of the useful statistical information leads to suboptimal results. This work aims to make the best use of the statistical information to improve the consistency between the given denoising results and the noise statistics, under the assumption that the noise is conditionally pixel-wise independent given the clean signal. A method, based on a Bayesian formulation of an auxiliary signal in the noisy data, is proposed for evaluating the consistency of the denoising results, without precise information on noise distribution. By leveraging the statistical information from noisy data, the method enhances the statistical noise consistency and improves denoising quality.
\end{abstract}


\section{Introduction} 

The modelling of image data and noise plays a central role in image processing. Real-world image acquisition processes often suffer from noise corruption, hence leading to varying degrees of degradation in the acquired data. 
An accurate understanding of the noise in the data is often a prerequisite for effectively handling many image processing tasks, such as denoising, image segmentation and image deblurring. 

There has been a substantial interest in developing noise removal schemes to obtain clean data in the past decades. At the heart of noise removal is the problem of decoupling two different signals from their combined version, and it usually requires understanding the characteristics of both the targeted signals and the noise. Many classic denoising algorithms such as Total Variation denoising \cite{rudin1992nonlinear},  non-local means \cite{buades2005non}, and BM3D \cite{dabov2007image} leverage fundamental image priors such as self-similarity \cite{buades2005non}, and sparsity in the gradient or wavelet domains \cite{coifman1995translation,donoho1995adapting}, to separate the image from noise and demonstrate remarkable success in recovering image details and mitigating noise artefacts. \added{PCA \cite{muresan2003adaptive, zhang2010two} and kPCA \cite{bakir2004learning} based methods achieve effective denoising by leveraging low-dimensional representations that capture local image structures.} In contrast, recent supervised deep learning-based denoising methods (e.g., \cite{chen2016trainable, zhang2017beyond, zhang2020residual}) do not explicitly model the images but learn denoising from a set of paired noisy-clean examples. These methods assume that sufficient paired noisy-clean examples exist which is not always the case in practice. Alternatively, unsupervised denoising methods \cite{lehtinen2018noise2noise, batson2019noise2self, metzler2018unsupervised, ke2021unsupervised} remove the need for clean data for training, by relying on assumptions on multiple noisy observations of each image  \cite{lehtinen2018noise2noise}, Gaussian white noise \cite{metzler2018unsupervised}, or zero-mean noise  \cite{batson2019noise2self, ke2021unsupervised}.

A key challenge for the application of the denoising methods is that the obtained results may be suboptimal if the noise distribution is unknown and the knowledge about the clean data is limited. The obtained denoising results might not align with the statistical characteristics of noise in different scenarios, for example, in classic model-based denoising methods where there is an error in the prior, or in recent unsupervised denoising learning algorithms where an inexact noise distribution assumption is used. Additionally, a direct evaluation of the consistency of the results with the noise characteristics is often not possible, because of the lack of calibration data or knowledge. One option to address this challenge is to reuse the noisy data and the encoded statistical information to study the consistency of the results. 

\begin{figure*}
    \centering
    \includegraphics[width=0.85\linewidth,trim={5 40 2 40}, clip]{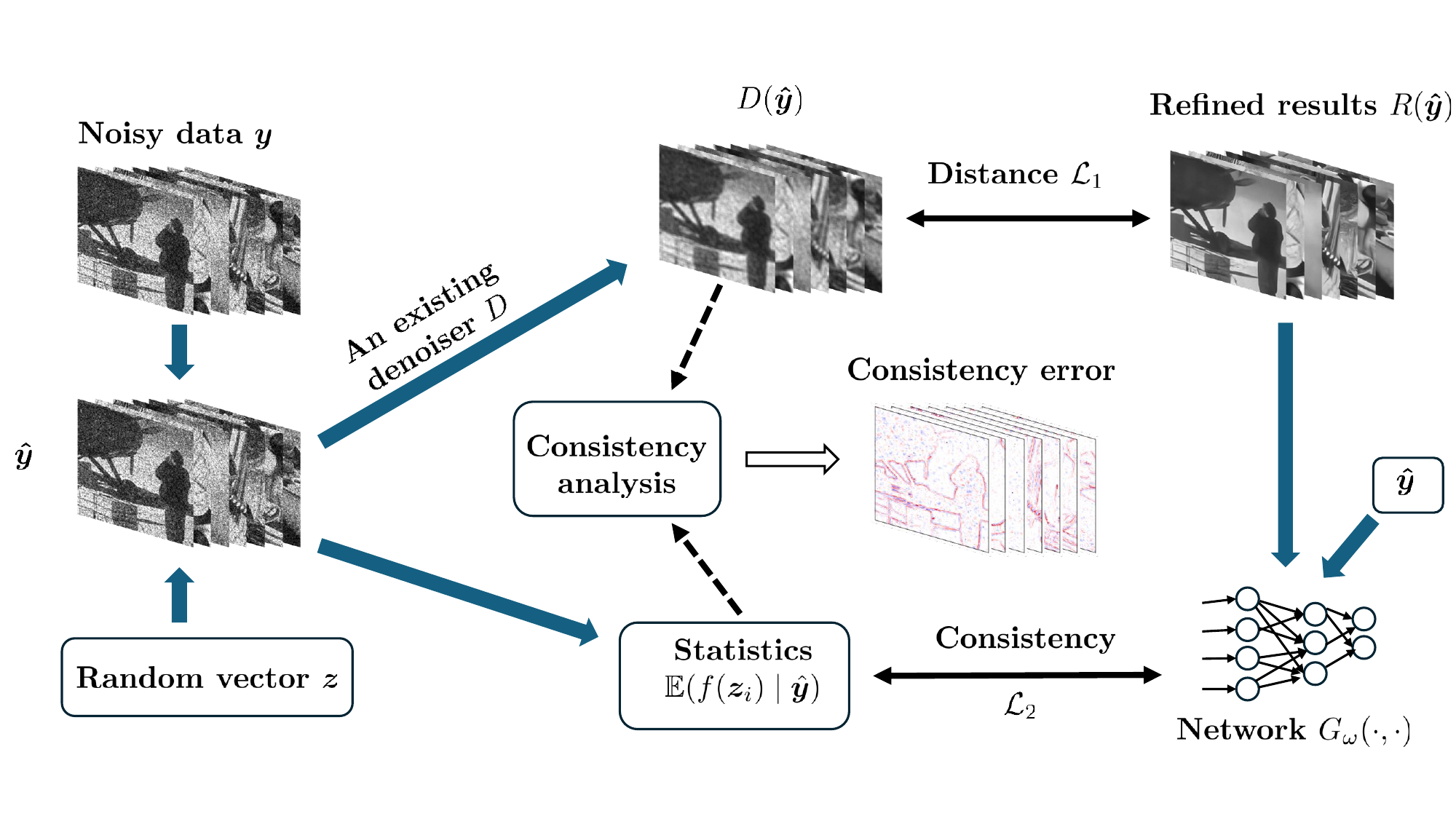}    
    \caption{The proposed methods for denoising refinement. Given a set of noisy data $\bmy$ and an existing denoiser $D$, the proposed method assesses the consistency between the results and the noise statistical information encoded in the data. Inconsistencies indicate the errors of the given denoising results. The results are refined by learning from the noisy data with a consistency constraint.}
    \label{fig:diagram}
\end{figure*}

In this work, we propose a method for evaluating and refining the results of an arbitrarily given denoising scheme, by learning from and exploiting the observed/available noisy data, under \textit{limited} knowledge and assumptions on the underlying noise distribution. The overall idea of the proposed method is illustrated in Figure \ref{fig:diagram}. 
The focus of this work is not on establishing new denoising methods, but we aim to improve the consistency between the results of an existing denoising method and the noise statistics. 
We assume that the noise is conditionally pixel-wise independent given the true signal. 
\added{This type of noise is commonly encountered in practice; examples include Poisson noise and Gaussian white noise. 
While spatially correlated noise is beyond the scope of this paper, preprocessing techniques such as whitening filters \cite{iqbal2018observation, kessy2018optimal, othman2021automated} offer a potential means of reducing correlations, allowing the noise to be treated in a manner similar to independent noise.
}

In the proposed approach, we incorporate an auxiliary signal (e.g., $\bmz$ in Figure \ref{fig:diagram}) into the data, which is then used to establish a criterion for analysing the consistency between the denoising results and the statistical characteristics of noise. 
The inconsistencies appearing at the different parts of the denoised data typically reflect systematic errors introduced by the existing denoising scheme (as illustrated by the red patterns in the centre of Figure \ref{fig:diagram}). To improve consistency, our method learns from a set of noisy data to find a better solution, with the help of the consistency criterion and a network $G_\omega$ (see the bottom right of Figure \ref{fig:diagram}). The experimental results show significant improvements over classic denoising schemes. 

\added{
Our method differs from recent unsupervised denoising approaches such as \cite{lehtinen2018noise2noise, krull2019noise2void, huang2022neighbor2neighbor} in that it is a refinement procedure rather than introducing a new denoising scheme. This feature makes it highly flexible, allowing it to be integrated with an existing denoising method to further exploit noisy observations. Moreover, many unsupervised denoising techniques rely on loss functions tailored to specific noise distributions (e.g., Gaussian noise \cite{soltanayev2018training}) or assume that the noise is zero-mean (e.g., \cite{krull2019noise2void, huang2022neighbor2neighbor, ke2024deep}). In contrast, our approach makes no such assumptions. We demonstrate that it can be applied to a variety of noise types, including non-zero-mean noise such as salt-and-pepper noise.}

The contributions of this work are summarised as follows. 
\begin{itemize}
    \item[1).] We propose a method for evaluating the consistency of denoising results without knowing the underlying noise distribution or having examples of noise-free data. The method works under the assumption that the underlying noise is pixel-wise independent given the clean signal.
    \item[2).] We develop a learning based refinement system that can improve the results of an existing denoising method, by reusing the statistical information from noisy data. 
    \item[3).] Our experimental results demonstrate the effectiveness of the proposed refinement approach in improving both the consistency and denoising quality of various classic approaches. 
\end{itemize}

\section{Related work}

The proposed learning framework works with existing denoising results and makes enhancements through the use of noisy data. Hence we review some existing works in denoising. We also review the related concepts of data consistency in parameter estimations.

\subsection{Model-based denoising and noise statistics}

There have been a variety of model-based denoising approaches. The accurate modeling of the image data and the noise is key to the success of these methods. One classic approach is total variation denoising \cite{rudin1992nonlinear}, which solves a variational problem to obtain the denoised result. Non-local means \cite{buades2005non} and BM3D \cite{dabov2007image}, on the other hand, leverage the redundancy of similar patches in the noisy image to construct the solution. The knowledge of the noise distribution is also a crucial factor for the performance of these methods. The LGSR method \cite{zha2022low} exploits the sparsity of similar patches within a group while simultaneously enforcing the low-rank structure of the group representation, thereby achieving effective denoising. 

Many of these methods work well for Gaussian noise. There are also approaches specifically designed to handle other types of noise, such as Poisson-Gaussian noise \cite{rakhshanfar2016estimation} and salt-and-pepper noise \cite{chan2005salt}. Additionally, these methods rely on predefined priors, which may not be optimal for all scenarios. In this work, we do not make any assumption about a specific noise distribution, but instead use instead use noise statistical information extracted directly from the noisy data to improve denoising performance. 

Noise statistics have been an integrated component of various denoising processes. The classic Wavelet Shrinkage methods (see e.g., \cite{donoho1995adapting, chipman1997adaptive}) rely on the estimated noise variance to compute the optimal thresholds of wavelet coefficients of the signals under the assumption of Gaussian noise. The estimated noise variances are also used to compute patch similarities in non-local means based approaches \cite{wu2013probabilistic, frosio2018statistical}. In PURE-LET \cite{luisier2010image}, the statistics of Poisson noise are used to derive an unbiased estimate of the expected
mean-squared error in the transformed domain. 
These methods typically use a summary statistic of the noise (such as an estimated noise variance) to determine hyper-parameters of underlying denoising models. However, this summary statistic does not account for all potential error patterns introduced by the denoising model itself—for example, when an inaccurate image prior is used.

\subsection{Deep learning methods} 
Deep learning has been a crucial part of recent advancements in denoising techniques. In particular, supervised deep-learning approaches have emerged as state-of-the-art methods for various denoising applications. Convolutional neural networks are a key element of these methods, for example, the U-net \cite{ronneberger2015u}, the Trainable Nonlinear Reaction Diffusion method \cite{chen2016trainable}, the DnCNN model \cite{zhang2017beyond}, and Swin-Conv-UNet \cite{zhang2023practical}.  

Unsupervised learning methods have also been developed for denoising. One of the first works in this domain is the Deep Image Prior (DIP) \cite{ulyanov2018deep}, which fits a randomly initialised network to noisy data and incorporates early stopping to avoid overfitting. MMES \cite{yokota2020manifold} is an alternative to DIP which models low-dimensional patch-manifold in embedded space of an autoencoder. By leveraging the prior, these methods are flexible and do not require ground truth training data. 

Noise2Noise \cite{lehtinen2018noise2noise} is another unsupervised approach that trains a denoising model using paired noisy observations. In the context of Gaussian noise, Stein's unbiased risk estimate (SURE) has been extended to learn denoisers from a set of unpaired noisy data \cite{soltanayev2018training}. Furthermore, a partially linear structure was introduced in \cite{ke2021unsupervised} to facilitate the learning of deep neural network denoisers from a collection of noisy images. Based on the assumption of zero-mean noise, a class of unsupervised approaches leverages the information of neighbouring pixels to design mean-squared-error (MSE) based objective functions creating blind spots in the input data. For example, Noise2Self \cite{batson2019noise2self} removes a subset of pixels and trains a network to predict them. Similarly to Noise2Nosie, Neighbor2Neighbor \cite{huang2022neighbor2neighbor} approximates the MSE loss using pairs of subsampled images. Noise2Fast \cite{lequyer2022fast} extends Neighbor2Neighbor by employing a customised downsampling technique to generate two noisy copies of an image from a single noisy image. 
Different from these methods, which construct a second noisy sample in the learning objective, our method leverages the statistical information of the noise encoded in the noisy data itself based on a predictive model. To our knowledge, this is the first work that proposes a learning-based framework to evaluate and refine existing denoising results, by using a predictive model to incorporate the statistical information from a set of noisy data.

\added{
Generative-model-based iterative refinement methods have recently gained attention for image restoration. Unconditional generative models, such as DDRM \cite{kawar2022denoising} and DreamClean \cite{xiao2024dreamclean}, use pretrained diffusion or score-based models as image priors to iteratively refine degraded inputs through alternating denoising and noise-injection steps. Iteratively preconditioned guidance \cite{garber2024image} further enhances convergence and fidelity. Although effective, these approaches depend on large networks preptrained on clean images and often assume a known corruption or degradation model. 
}

\added{
Conditional generative models extend this idea by conditioning refinement on the noisy input or an intermediate estimate. Palette \cite{saharia2022image} showed the effectiveness of conditional diffusion models for image restoration. Predict-and-refine frameworks \cite{whang2022deblurring} and transformer-based iterative denoising methods \cite{xu2024diversified} jointly learn deterministic predictors and stochastic refiners for efficient restoration. Broader conditional diffusion and iterative refinement frameworks \cite{pandey2024fast} achieve strong results while preserving image-specific structures. In contrast, our method directly models unknown noise structures using noisy data and leverages the extracted information for refinement. It does not rely on pretrained generative models or clean ground truth data. 
}

\added{
Denoising autoencoders (DAEs) \cite{vincent2008extracting, im2017denoising} are a class of methods that learn representations by introducing artificial noise into the input data during training. The added noise defines a denoising objective, which can be combined with additional regularisation terms \cite{lee2021noise}. However, the role of the added noise in DAEs is primarily to encourage the encoder to learn robust low-dimensional representations rather than to explicitly remove noise from the data.
}

\subsection{Data consistency in parameter estimations.} 
Data consistency is a key criterion for denoising, inverse problems, and many other parameter estimation tasks.  It ensures that the solutions accurately reproduce the provided data. Various approaches for inverse problems impose hard constraints on data consistency on the solutions. In particular, one important class of these approaches is the variational methods. These methods typically seek solutions by minimising an objective function (see e.g., \cite{rudin1992nonlinear, bredies2010total}), which explicitly includes regularisation terms and data-fitting terms. The latter penalises the distance between the solutions and the measurements (noisy observations) in the measurement domain. The proximity between the solutions and the data does not necessarily lead to good results, particularly in the presence of noise. For example, if there is an error in the prior (regularisation), it can lead to structured errors in the denoised results while the data fitting term remains small. There is an essential difference between these approaches and our method, which focuses on consistency from a statistical perspective. Importantly, we explore the consistency with noise statistics to fix the errors caused by the reconstruction models themselves, which can not be done by simply penalising the distance between the solutions and the measurements.

\section{Statistical refinement for denoising}

Let $\bmx$ be a random vector in $\bbR^n$ representing an image, and $\bmy \in \bbR^n$ be a random vector representing an observed noisy version of $\bmx$. 
We define 
\begin{equation} \label{eq:noise}
\bmn := \bmy - \bmx    
\end{equation}
which represents the noise in the observed data. It is worth mentioning that the expression in \eqref{eq:noise} generally defines the noise as the difference between $\bmy$ and $\bmx$, and this does not indicate that the noise is additive. Given \eqref{eq:noise}, the task of denoising involves obtaining an estimate of $\bmx$ given the data $\bmy$.

For the simplicity of notations, we assume throughout this section that $\bmx$ and $\bmn$ are both continuous random variables but the results also hold for the discrete cases. We denote the probability density of a random vector $\bm{a}$ by $p(\bm{a})$ and the conditional density of $\bm{a}$ given $\bm{b}$ by $p(\bm{a} \mid \bm{b})$. The notation $\bbE\qut{\bm{a} \mid \bm{b}}$ denotes the conditional expectation of $\bm{a}$ given $\bm{b}$.

In this paper, we assume that the noise $\bmn$ is pixel-wise independent conditioned on the image $\bmx$. 
More formally, letting $\bmn_i$ and $\bmx_i$ denote the $i^{\rm th}$ element of $\bmn$ and $\bmx$, respectively, and letting $\bmn_i^c \in \bbR^{n-1}$ (resp. $\bmx_i^c$ or $\bmy_i^c$) denote the vector containing the remaining $n-1$ elements of $\bmn$ (resp. $\bmx$ or $\bmy$), we have the following assumption on $\bmn$. 
\begin{Assumption}\label{Assumption1}
For any $i$, it holds that the probability density $p(\bmn_i \mid \bmx_i, \bmn_i^c) = p(\bmn_i \mid \bmx_i)$.
That is, $\bmn_i$ is independent of $\bmn_i^c$ conditioned on $\bmx_i$.
\end{Assumption}

As a consequence of the above assumption, it follows that the noisy pixels $\bmy_i$ and $\bmy_j$ are also conditionally independent given $\bmx_i$ for all $j \neq i$. 
However, it is important to note that we do not assume that the specific distribution of $\bmn$ is known, or that $\bmn$ is independent of $\bmx$. Therefore the noise can be signal-dependent.

\subsection{Statistical noise consistency}

Given the noisy data $\bmy$, the aim of the denoising task is to obtain an estimate of $\bmx$ from $\bmy$. A good estimate of $\bmx$ should recover as much information of $\bmx$ as possible. An ideal approach to do this is to estimate the posterior density of $\bmx$ given $\bmy$, denoted by
\[
p(\bmx \mid \bmy).
\]
This is equivalent to obtaining the conditional distribution of the noise $p(\bmn \mid \bmy)$ according to the relationship \eqref{eq:noise}. 

In practical situations, however, computing the posterior $p(\bmx \mid \bmy)$ can be challenging, especially when samples of the paired data ($\bmy$, $\bmx$) are not readily available. One possible way of estimating the posterior is to use the Bayesian theorem $p\qut{\bmx \mid \bmy} = {p\qut{\bmy \mid \bmx} \cdot p(\bmx)} / {p(\bmy)}$, provided that the noise model $p\qut{\bmy \mid \bmx}$ and the prior $p(\bmx)$ are known or can be estimated. However, one of the main obstacles here is that neither the noise model nor the samples of $\bmx$ are given. 

In this work, we are interested in answering the question of \textit{whether a given estimate $\hat{\bmx}$ of $\bmx$ is consistent with the statistics of noise in ${\bmy}$}. The denoising outcome should have the ability to reproduce certain characteristics of the noise within the noisy data. In particular, $\hat{\bmx}$ should be aligned with statistical estimators of the noise which can be obtained given $p(\bmn \mid \bmy)$. However, it is challenging to verify this property of $\bmxh$ if $p(\bmn \mid \bmy)$ is unknown. 

One motivation for the above question comes from the fact that the estimates $\bmxh$ may not be consistent with the statistics of the noise when they are obtained without accurate knowledge about the probability distributions of the image and noises. Improper estimation may lead to a substantial loss of information from the data and hence unsatisfactory denoising outcomes. In particular, if there are unexpected artefacts or missing image structures in the denoised image $\hat{\bmx}$, the estimated noise $\hat{\bmn}:=\bmy - \hat{\bmx}$ is also corrupted by such errors and hence does not match the true noise statistics. 

In the following, we will explore the consistency of the denoising outcomes and the noise statistics without knowing the distribution of noise or samples of clean data.  We will introduce statistical noise consistency based on an auxiliary random signal in the data, and then propose a learning method to leverage the noisy data to impose consistency and improve the denoising outcomes. 

We introduce a random auxiliary signal, denoted as $\bmz \in \bbR^{n}$, and we define 
\begin{equation}\label{eq:yhat}
    \bmyh := \bmy + \bmz,
\end{equation}
where $\bmz$ satisfies the following assumption. 

\begin{Assumption}\label{Assumption2}
For any $i$, the conditional probability of $\bmz_i$ satisfies $p(\bmz_i \mid \bmy_i, \bmy_i^c, \bmz_i^c, \bmx) = p(\bmz_i \mid \bmy_i)$. 
\end{Assumption}

This assumption implies that $\bmz_i$ does not depend on $\bmz_i^c$, $\bmx$, and $\bmy_i^c$ if $\bmy_i$ is observed. Here $\bmz$ is considered to be a known signal and the variance of $\bmz$ is much smaller than that of the noise $\bmn$. With $\bmyh$ being an alteration of original data $\bmy$, we can explore how to estimate $\bmx$ from $\bmyh$.

While the noise $\bmn$ is unknown given $\bmyh$, the auxiliary vector $\bmz$ is given. Instead of working directly on $\bmn$, we focus on $\bmz$. Assuming that we have an estimate of the posterior distribution of $\bmx$ given $\bmyh$, it should provide useful information for estimating $\bmz$. To exploit this property, we define the following concept of consistency in estimation. 
\begin{Definition*}[Statistical noise consistency] 
Suppose that we have an estimate of the conditional distribution of $\bmx_i$ given $\bmyh$, denoted by $p_{\theta, \bmyh, i}(\bmxh_i)$, for $i=1,2,\cdots,n$. The distribution $p_{\theta, \bmyh, i}(\bmxh_i)$ is said to be consistent if, for any given Borel measurable function $f$, there exists a mapping $\mathcal{M}$ which is independent of $i$ and satisfies that 
    \begin{equation}\label{eq:consistency}
        \bbE (f(\bmz_i) \mid \bmyh ) = \mathcal{M} (p_{\theta, \bmyh, i}(\cdot), \bmyh_i), \quad i=1,2,\cdots, n,
    \end{equation}
    where the expectation $\bbE(f(\bmz_i) \mid \bmyh )$ is taken over $\bmx$ and $\bmz_i$ given $\bmyh$. 
\end{Definition*}

In the above definition, the density function $p_{\theta, \bmyh, i}(\bmxh_i)$ and $\bmyh_i$ retain all information from $\bmyh$ useful for inferring the posterior of $\bmz_i$ given $\bmyh$. In other words, if we have $p_{\theta, \bmyh, i}(\bmxh_i)$ and $\bmyh_i$, we no longer require $\bmyh_i^c \in \bbR^{n-1}$ for predicting $\bmz_i$. 
This property does not hold generally if the estimated $\bmxh_i$ contains unexpected error patterns (such as over-smoothed image structures). In this case, $p_{\theta, \bmyh, i}(\bmxh_i)$ is no longer sufficient for recovering $\bbE (f(\bmz_i) \mid \bmyh )$ given only $\bmyh_i$. 

The following proposition provides a justification for using \eqref{eq:consistency} as a criterion for checking the consistency of denoising results. We will show that\footnote{\added{A similar result for correlated noise, where there is a close relationship between the expectation $\bbE \qut{ f(\bmz_i) \mid  \bmyh }$ and the uncorrelated component of the noise, is given in Proposition 2 in the Appendix.}}, under Assumptions 1 and 2, the true posterior $p(\bmx_i \mid \bmyh)$ satisfies the condition \eqref{eq:consistency}.  

\begin{Proposition}\label{prop:1}
    Let $\bmyh$ be defined in \eqref{eq:yhat} and $f$ be a Borel measurable function. Under Assumptions \ref{Assumption1} and \ref{Assumption2}, there exists a two dimensional function $G_\omega: \bbR^2 \rightarrow \bbR$, such that, for given $\bmyh$, the conditional expectation of $f(\bmz_i)$ can be expressed as  
    \begin{equation}\label{eq:fz}
        \bbE \qut{ f(\bmz_i) \mid  \bmyh } = \int p\qut{\bmx_i \mid \bmyh} G_\omega(\bmx_i, \bmyh_i) d \bmx_i.
    \end{equation} 
    Consequently, to compute the conditional expectation $\bbE \qut{ f(\bmz_i) \mid  \bmyh }$, it is sufficient to know the posterior $p\qut{\bmx_i \mid \bmyh}$, $\bmyh_i$, and a two dimensional function $G_\omega$. Here, the function $G_\omega$ may depend on $f$.
\end{Proposition}
\begin{proofsketch}
The above statement follows from an application of the law of total expectation. Considering $\bmyh$ as a fixed signal, by the law of total expectation, we have
\begin{align*}
\bbE\qut{ f(\bmz_i) \mid  \bmyh } & = \bbE_{\bmx_i}\qut{\bbE\qut{ f(\bmz_i) \mid \bmx_i, \bmyh }} \\
& = \int p(\bmx_i \mid \bmyh )  \cdot \bbE\qut{ f(\bmz_i) \mid \bmx_i, \bmyh } d \bmx_i,
\end{align*}
where $\bbE_{\bmx_i}$ represents the expectation taken over $\bmx_i$ (here given the $\bmyh$), and $p(\bmx_i \mid \bmyh )$ denotes the conditional probability density of $\bmx_i$ given $\bmyh$. 

If we let $G_\omega(\bmx_i, \bmyh_i):=\bbE\qut{ f(\bmz_i) \mid \bmx_i, \bmyh_i }$, it remains to show that $\bbE\qut{ f(\bmz_i) \mid \bmx_i, \bmyh }=\bbE\qut{ f(\bmz_i) \mid \bmx_i, \bmyh_i }$. This is a consequence of Assumptions \ref{Assumption1} and \ref{Assumption2}. Please refer to the Appendix for details. 
\end{proofsketch}

Equation \eqref{eq:fz} implies that the true posterior $p\qut{\bmx_i \mid \bmyh}$ satisfies the consistency condition in \eqref{eq:consistency}, if we let $\mathcal{M}(p\qut{\cdot \mid \bmyh}, \bmyh_i):=\int p\qut{\bmx_i \mid \bmyh} G_\omega(\bmx_i, \bmyh_i) d \bmx_i$. If $p_{\theta, \bmyh, i}(\cdot)$ is equivalent to $p\qut{\bmx_i \mid \bmyh}$ (i.e., an exact estimate of the posterior),  \eqref{eq:fz} can be rewritten as
\begin{equation}\label{eq:fz2}
\bbE \qut{ f(\bmz_i) \mid  \bmyh } = \langle p_{\theta, \bmyh, i}(\cdot),  G_\omega(\cdot, \bmyh_i) \rangle,  
\end{equation}
where $\langle \phi_1, \phi_2 \rangle := \int \phi_1(x) \phi_2(x) d x$ denotes the inner product of two real-valued functions $\phi_1$ and $\phi_2$. Therefore, $\bbE \qut{ f(\bmz_i) \mid  \bmyh }$ can be computed as the inner product of two functions: the posterior $p_{\theta, \bmyh, i}(\cdot)$ and the function $G_\omega(\cdot, \bmyh_i)$. It is worth mentioning that, while $\bmz$ is a known signal, the function $G_\omega$ is dependent on the noise distribution and hence unknown.

\begin{figure}[ht!]
    \centering
    \footnotesize 
    \begin{tabular}{c} 
    \includegraphics[width=0.9\linewidth,trim={0 5 0 0}, clip]{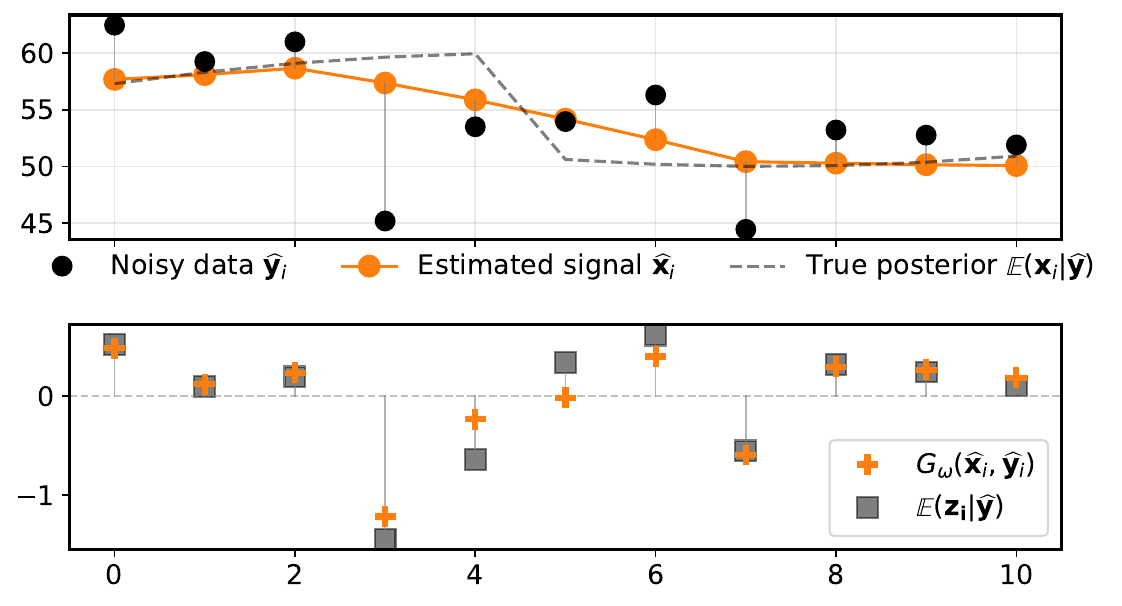} \\
    \added{(a). Original estimated signal (top) and  $G_{\omega}\qut{\bmxh_i, \bmyh_i}$ (bottom)}\\
    \includegraphics[width=0.9\linewidth,trim={0 5 0 0}, clip]{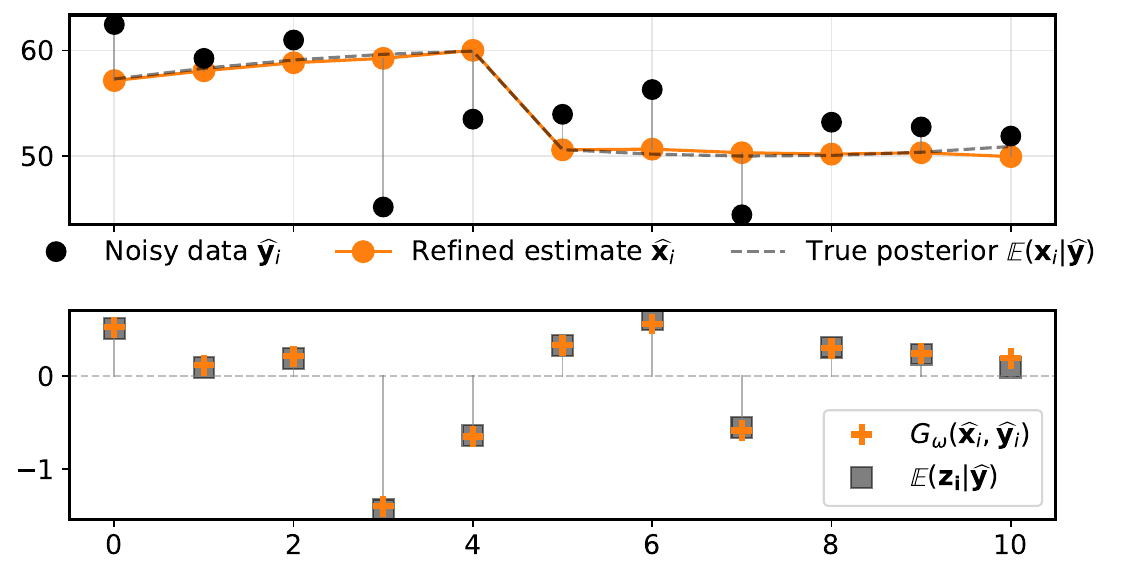} \\
    \added{(b). Refined signal (top) and  better $G_{\omega}\qut{\bmxh_i, \bmyh_i}$ (bottom)}\\
    \end{tabular}
    \caption{\added{Toy example: Refining estimates using $G_\omega$ as a constraint.}}   \label{fig:toy_example}
\end{figure}

\added{\textbf{\textit{Toy example. }}} 
\added{
Let $\bmn_i \sim \mathcal{N}(0,6^2)$ and $\bmz_i \sim \mathcal{N}(0,2^2)$ for all $i$, and let $f(x)=x$. Then
$
\mathbb{E}(\bmz_i | \bmyh) 
= \int p(\bmx_i | \bmyh) \, \mathbb{E}[\bmz_i | \bmx_i, \bmyh] \, d\bmx_i
= \int p(\bmx_i | \bmyh) \, \frac{2^2}{2^2 + 6^2} (\bmyh_i - \bmx_i) \, d\bmx_i
= 0.1 \, (\bmyh_i - \mathbb{E}[\bmx_i | \bmyh]).
$
Define linear function $G_{\omega}(x, \bmyh_i) := 0.1 \, (\bmyh_i - x)$. For simplicity, assume the estimated posterior is a Dirac distribution $p_{\theta, \bmyh, i} := \delta_{\bmxh_i}$ at the estimated signal $\bmxh_i$, which means that the constraint in (6) becomes
$
\mathbb{E}(\bmz_i | \bmyh) = G_{\omega}(\bmxh_i, \bmyh_i).
$
If $\bmxh_i:=D_i(\bmyh)$ deviates substantially from the posterior mean $\mathbb{E}[\bmx_i | \bmyh_i]$, $G_{\omega}(\bmxh_i, \bmyh_i)$ differs significantly from $\mathbb{E}(\bmz_i | \bmyh)$, violating the constraint (see Figure~\ref{fig:toy_example}(a)). Solving (6) to adjust $\bmxh_i$ so that $\mathbb{E}(\bmz_i \mid \bmyh) \approx G_{\omega}(\bmxh_i, \bmyh_i)$ enforces an estimate to be closer to the posterior mean (see Figure~\ref{fig:toy_example}(b)).
}

\textbf{\textit{Remark 1.}} In \eqref{eq:fz2}, $G_\omega(\cdot, \bmyh_i)$ depends on $\bmyh_i$ but not on the location $i$. This property does not hold if we replace the second argument of $G_\omega$ by $\bmyh$, hence breaking the assumption of $\mathcal{M}$ in criterion \eqref{eq:consistency}. Such a replacement makes $G_\omega$ dependent on $i$ and is generally not helpful for studying $p_{\theta, \bmyh, i}(\cdot)$, as it includes a special case $G_\omega(\bmx_i, \bmyh) := \bbE \qut{ f(\bmz_i) \mid  \bmyh }$, in which \eqref{eq:fz2} holds for any $p_{\theta, \bmyh, i}(\cdot)$. 

In summary, Proposition \ref{prop:1} and \eqref{eq:fz2} establish a connection between the posterior of $\bmx_i$ and the auxiliary signal $\bmz$. This connection can be leveraged as a criterion to assess the statistical noise consistency of predicted denoising solutions from the data, with the help of a two-dimensional function $G_\omega$. In the next part of this section, we will present a learning based approach to make use of this criterion for enhancing the consistency and denoising quality.

\subsection{Learning-based denoising refinement}\label{subsect:refinement}

We propose an approach to refine the denoising results using \eqref{eq:fz2}. Given an initial estimate of $\bmx$, our objective is to obtain the refined versions, represented by an estimated posterior $p_{\theta, \bmyh, i}(\bmxh_i)$ where $\theta$ is the parameter of the model, that satisfy the condition \eqref{eq:fz2}. Since $G_\omega$ is unknown, we model it as a parameterised function with parameter $\omega$ which can be learnt from the data.

Suppose that a denoiser, denoted by $D(\bmyh)$, is given. We want to find $\bmxh$ from the neighbourhood of $D(\bmyh)$ such that it satisfies \eqref{eq:fz2}. This motivates the following optimisation problem for $\theta$ and $\omega$.
\begin{equation}\label{eq:constrainedopt}
\begin{aligned}
\min_{\theta, \omega}  \ &  \bbE_{\bmyh} \qut{ \sum_{i=1}^n \bbE_{\bmxh_i  \sim  p_{\theta, \bmyh, i}(\cdot) } ( D_i(\bmyh)  - \bmxh_i )^2} \\
\text{subject to } &  \bbE \qut{ f(\bmz_i) \mid  \bmyh } \!=\! \left\langle p_{\theta, \bmyh, i}(\cdot), G_{\omega}\qut{\cdot, \bmyh_i} \right\rangle, \ i\!=\!1,\!\cdots,\!n,
\end{aligned}    
\end{equation}
where $D_i(\bmyh)$ denotes the $i^{\rm th}$ entry of $D(\bmyh)$, and $\bbE_{\bmxh_i  \sim  p_{\theta, \bmyh, i}(\cdot) }$ represents the expectation taken over the distribution of $p_{\theta, \bmyh, i}(\cdot)$.
\added{Here, $\bmxh_i$ is distributed according to $p_{\theta, \bmyh, i}(\cdot)$ and therefore depends on both $\theta$ and $\bmyh$.}
The constraint in \eqref{eq:constrainedopt} imposes the condition \eqref{eq:fz2} on the solution. 

\textbf{\textit{Remark 2. }} The functions $p_{\theta, \bmyh, i}(\cdot)$ and $G_{\omega}\qut{\cdot, \bmyh_i}$ that satisfy the constraint in \eqref{eq:constrainedopt} are not unique. To see this point, if they satisfy the constraint, then the functions $p_{\theta, \bmyh, i}(\cdot - t)$ and $G_{\omega}\qut{\cdot-t, \bmyh_i}$ also satisfy the constraint for any $t \in \bbR$. In other words, having $p_{\theta, \bmyh, i}(\cdot) =p(\bmx_i \mid \bmyh)$ is sufficient but not necessary for computing $\bbE \qut{ f(\bmz_i) \mid  \bmyh }$. Nevertheless, the constraint enhances the consistency of the solutions.

\added{\textit{B.1 Approximating the inner product.}}
In practical computations, we replace the inner product in \eqref{eq:constrainedopt} with a finite sum.
We model the distribution $p_{\theta, \bmyh, i}(\cdot)$ with a finite set of samples $\{R_{\theta,k,i}
(\bmyh)\}_{k=1,\cdots,K}$ where $R_{\theta,k,i}$ is a function parameterised by $\theta$, and $K$ is a large number. Given that $\{R_{\theta,k,i}
(\bmyh)\}_{k=1,\cdots,K}$ represent a set of samples drawn from $p_{\theta, \bmyh, i}(\cdot)$, the inner product is approximated by the following finite sum
\[
\left\langle p_{\theta, \bmyh, i}(\cdot), G_{\omega}\qut{\cdot, \bmyh_i} \right\rangle \approx \frac{1}{K} \sum_{k=1,...,K} G_{\omega}\qut{ R_{\theta,k,i}(\bmyh), \bmyh_i }.
\]
Accordingly, we have the following approximation of the expectation 
\[
\bbE_{\bmxh_i  \sim  p_{\theta, \bmyh, i}(\cdot) } ( D(\bmyh)_i  - \bmxh_i )^2 \approx \frac{1}{K} \sum_{k=1,...,K} ( D_i(\bmyh)  - R_{\theta,i,k}(\bmyh) )^2. 
\]

\begin{figure}[ht!]
    \centering
    \includegraphics[width=\linewidth, trim={150 222 158 160}, clip]{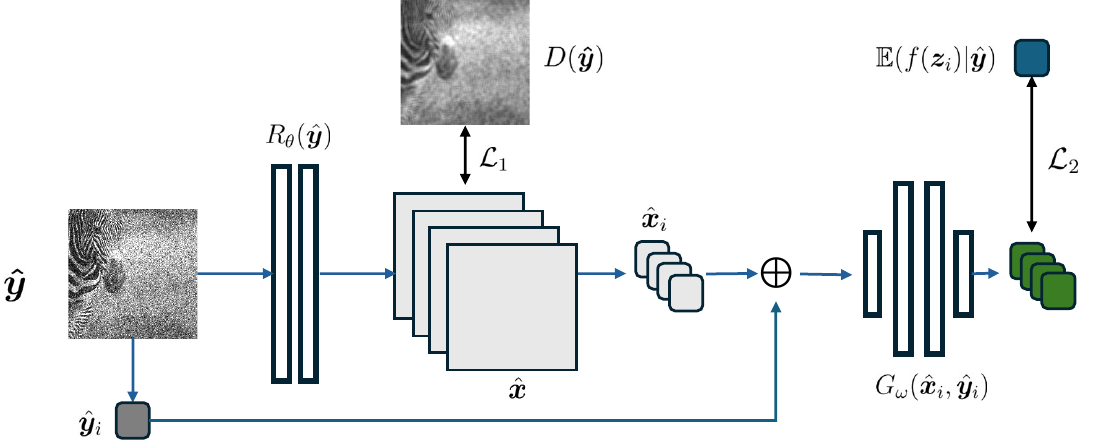}
    \caption{\added{Illustration of the proposed learning method}}
    \label{fig:learning_system}
\end{figure}

For simplicity, we use $R_{\theta,k}(\bmyh)$ to denote the vector of $R_{\theta,k,i}(\bmyh)$ ($i=1,\cdots,n$) and similarly use $G_{\omega}\qut{R_{\theta,k}(\bmyh), \bmyh}$ to denote the vector of $G_{\omega}\qut{ R_{\theta,k,i}(\bmyh), \bmyh_i }$ ($i=1,\cdots,n$). 
The constrained optimisation problem \eqref{eq:constrainedopt} can be addressed using a penalty method, resulting in the formulation:
\begin{equation}\label{eq:unconstrainedopt}
\begin{aligned}
\min_{\theta, \omega}  \  \bbE_{\bmyh} \Bigg[  & \underbrace{\frac{1}{K} \sum_{k=1}^K \| D(\bmyh)  - R_{\theta,k}(\bmyh) \|_2^2}_{:= \mathcal{L}_1(\bmyh)}   \\
& + \underbrace{ \lambda \mathcal{D}\qut{ {\frac{1}{K}}\sum_{k=1}^K G_{\omega}\qut{ R_{\theta,k}(\bmyh), \bmyh }, \ \ \bbE \qut{ f(\bmz) \mid  \bmyh } } }_{:= \mathcal{L}_2(\bmyh)} \Bigg] \\
\end{aligned}    
\end{equation}
where $\lambda$ is the penalty coefficient, $\mathcal{D}\qut{\cdot, \cdot}$ is a distance metric. The second term $\mathcal{L}_2(\bmyh)$ enforces the constraint of problem \eqref{eq:constrainedopt}. In the context of deep learning, both $R_{\theta,k}$ and $G_{\omega}$ can be implemented using deep neural networks. 
\added{The refinement method is illustrated in Figure \ref{fig:learning_system}}.

\added{\textit{B.2 multiple consistency criteria.}}
The constraint in \eqref{eq:fz2} can be extended into multiple criteria if multiple versions of the function $f$ are used. Assuming that we have $L$ functions $\{ f_1, \cdots, f_L\}$, the associated loss function is given by
\begin{equation}\label{eq:L2-2}
 \mathcal{L}(\bmyh)  := \mathcal{L}_1(\bmyh) + \mathcal{L}_2(\bmyh) 
\end{equation}
where $\mathcal{L}_1$ is given in \eqref{eq:unconstrainedopt}, and 
\begin{equation*}
\begin{split}
\mathcal{L}_2(\bmyh) & :=  
 \frac{\lambda}{L} \sum_{l=1}^L \mathcal{D}  \left( { \frac{1}{K}}\sum_{k=1}^K G_{\omega, l}\qut{ R_{\theta,k}(\bmyh), \bmyh }, \bbE \qut{ f_l(\bmz) \mid  \bmyh } \right).
\end{split}
\end{equation*} 

\added{Once the model $R_{\theta,k}(\bmyh)$ is trained, denoised results for unseen noisy images can be obtained by computing the average $\frac{1}{K} \sum_{k=1}^KR_{\theta,k}(\bmy)$, without needing to compute $\hat{\bmy}$. The motivation behind this is that $\bmy$ is closer to the clean image $\bmx$, making it inherently easier to denoise.}

\subsection{The auxiliary signal} 

While Proposition \ref{prop:1} holds for any $\bmz$ that satisfies Assumption \ref{Assumption2}, in practice $\bmz$ should be chosen to avoid the problem \eqref{eq:fz2} admitting trivial solutions for $p_{\theta, \bmyh, i}(\cdot)$. To illustrate this, suppose the value $\bmy_i$ is taken from a discrete set ${1, 2, 3, \cdots}$, and let $\bmz_i$ be drawn from a uniform distribution on $[-0.1, 0.1]$. Then, $p_{\theta, \bmyh, i}(t) := 1$ and $G_\omega(t, \bmyh_i) := \bmy_i - \mathrm{round}(\bmyh_i)$ for any $t$ is a trivial solution to \eqref{eq:fz2}, where $\mathrm{round}(\cdot)$ rounds to the nearest integer. This is because $\mathrm{round}(\bmyh_i) = \bmy_i$, and
\[
\begin{split}
\bbE \qut{ f(\bmz_i) \mid  \bmyh } 
 = f(\bmz_i)  
 & = f(\bmy_i - \mathrm{round}(\bmyh_i)) \\
 & = \langle 1,  f(\bmy_i - \mathrm{round}(\bmyh_i)) \rangle.  
\end{split}
\]

In general, the choice of $\bmz$ should satisfy the following two conditions: 
\begin{enumerate} 
\item The value of $\bmz_i$ cannot be obtained from $\bmy_i + \bmz_i$ alone. 
\item The variance of $\bmz_i$ is small compared to the noise variance. 
\end{enumerate} 
The second condition ensures that $\bmyh \approx \bmy$, since otherwise $R_{\theta,k}(\bmyh)$ may not effectively remove noise from $\bmy$ due to the elevated noise level in $\bmyh$.

Taking these conditions into consideration, a practical example for $\bmz$ is as follows. We define
\begin{equation}\label{eq:z_example}
    \bmz:=\bmyh-\bmy, \ \text{where} \ \bmyh:= {\rm discr}(\bmy + M \circ \bm{r}),
\end{equation}
in which $M$ is a random binary mask, $\bm{r}$ is a vector of i.i.d. random variables (for example, Gaussians), $\circ$ denotes element-wise multiplication, and the operator $\mathrm{discr}$ denotes the discretisation operation that rounds each element of a vector to the nearest pixel intensity level from $\bmy$ if $\bmy$ has discrete values, or acts as the identity function otherwise. Here, the mask $M$ can be chosen to be sparse, so that only a small portion of elements in $\bmy$ are perturbed to ensure $\bmyh \approx \bmy$ as required by the condition 2). The operator $\mathrm{discr}$ ensures consistency in the variable type of $\bmyh$, preventing the mixing of continuous and discrete values when $\bmy$ is discrete. This is motivated by condition 1) because it is relatively easier to infer $\bmz_i$ if $\bmz_i$ is continuous while $\bmy_i$ is discrete.

\subsection{Optimisation} 
Given a set of samples $\{\hat{y}^{(j)}\}$ of the noisy data $\bmyh$, one can optimise \eqref{eq:L2-2} using min-batch optimisation based algorithms such as stochastic gradient descents (SGD). During the optimisation process, it is desirable to choose a large $\lambda$ to enforce the constraint in \eqref{eq:constrainedopt}. 
However, one main challenge is that the iterations are prone to get trapped in local minima of $G_{\omega}$, if the gradient of $\mathcal{L}_2$ with respect to $\theta$ is much bigger than that of $\mathcal{L}_1$. According to the chain rule, the gradient $\partial \mathcal{L}_2 / \partial \theta = \partial \mathcal{L}_2 / \partial R_{\theta,k}(\hat{y})  \cdot  \partial R_{\theta,k}(\hat{y}) / \partial \theta $. Even with a small value of $\lambda$, the gradient $\partial \mathcal{L}_2 / \partial \theta$ can still be large for some $\hat{y}$ as the first term $\partial \mathcal{L}_2 / \partial R_{\theta,k}(\hat{y})$ in the product depends on the structure of $G_{\omega}$. 

To prevent very large gradients of $\partial \mathcal{L}_2$ with respect to $\theta$, we rescale $\partial \mathcal{L}_2 / \partial R_{\theta,k}(\hat{y})$ during the gradient computation process. Given a sample $\hat{y} \in \{\hat{y}^{(j)}\}$. Let $g_1 := \partial \mathcal{L}_1 / \partial R_{\theta,k}(\hat{y}) $ and $g_2 := \partial \mathcal{L}_2 / \partial R_{\theta,k}(\hat{y})$. We define
\[
g := \frac{g_1 + \gamma\cdot( s(g_2) + s({\rm sgn}(g_2) )/2 }{1+\gamma},
\]
where ${\rm sgn}(g_2)$ is the element-wise sign function, $s(\cdot)$ is a function defined by $s(a)= (\|g_1\|_2/\|a\|_2) a $ for $a \in \bbR^n$, which implies $\|s(a)\|_2 = \|g_1\|_2$. %
If $\gamma=0$, then $g=g_1$ and hence $\partial \mathcal{L}_2 / \partial R_{\theta,k}(\hat{y})$ is ignored.
The denominator ${1+\gamma}$ prevents very large $g$ when $\gamma$ is large. It is worth mentioning that the size of $g$ does not depend on $\lambda$. Finally, the gradient of $\theta$ is computed by
$
g \cdot (\partial R_{\theta,k}(\hat{y}) / \partial \theta).
$
Note that the gradient for $\omega$ is unchanged. During the optimisation process, $\gamma$ is chosen to be small initially to emphasise $g_1$, and then it is gradually increased.

\section{Experimental results
}\label{sec:exp}

The goal of the experiments is to evaluate the effectiveness of the refinement approach in improving the consistency of the denoised results and the quality of different denoising schemes. We will compare the statistical noise consistency of the results and the performance gap between the refined results and supervised learning baselines, for different types of noise. All experiments are carried out on a computing device equiped with a NVIDIA Tesla P100 GPU (16 GB) and 8 Intel(R) Xeon(R) CPUs (2.40GHz each). 
The proposed method\footnote{code:  https://github.com/RK621/Statistical-refinement-for-denoising} is implemented using PyTorch and tested using benchmark denoising datasets. 

\vspace{0.2cm}
\noindent\textbf{Datasets.} Our experiments utilise a benchmark denoising dataset consisting of $400$ images with dimensions of $180 \times 180$ pixels, following the setting of \cite{chen2016trainable, zhang2017beyond}. The training is performed using noisy versions of these $400$ images (the ground truth images are not used). The quality of the denoising results is evaluated and compared on the widely-used test datasets BSD68 \cite{martin2001database} and Set12 \cite{zhang2017beyond}. 
We consider different types of noise at different noise levels: 1) Gaussian noise, 2) Poisson noise which is zero-mean and signal-dependent, 3) salt-and-pepper noise \cite{bovik2010handbook} which is not zero-mean, and 4) mixed Poisson and Gaussian noise.  

\vspace{0.2cm}
\noindent\textbf{Network architectures.}
Our method is agnostic to network architectures. 
To ensure fair comparison over different scenarios, we employ the same network architecture for $\{R_{\theta, k}\}_{k=1}^K$ (being used in Equation \eqref{eq:L2-2}) throughout the experiments, which is derived from the benchmark denoising model DnCNN \cite{zhang2017beyond}. $\{R_{\theta, k}\}_{k=1}^K$ are obtained by modifying the last layer of DnCNN to output $K$ images, where $K=64$. The network $\{G_{\omega,l}\}_{l=1}^L$ (being used in Equation \eqref{eq:L2-2}) consists of $6$ convolutional layers with a kernel size of $1$ and includes residual connections. The choice of a kernel size of $1$ facilitates element-wise operations of $G_{\omega,l}(\cdot, \cdot)$, ensuring that the $i^{\rm th}$ element of its output vector solely depends on the corresponding $i^{\rm th}$ elements of the input vectors (as required by condition \eqref{eq:fz2}). Please refer to the supplemental material for further details of the architectures of $R_\theta$ and $G_\omega$.

\vspace{0.2cm}
\noindent\textbf{Generating samples of $\bmyh$.}
The samples of $\bmyh$ are generated using \eqref{eq:z_example}, where we combine samples of noisy images and randomly generated $\bmz$ using $\bm{r}$. In our implementation, the $\bm{r}$ is a vector of i.i.d. Gaussians with zero means. The random mask $M$ is chosen such that it has an average of $1/256$ nonzero entries for salt-and-pepper noise and $1/64$ nonzero entries for other noise. The standard deviation of the entries of $\bm{r}$ is set to $0.05$ for Gaussian, Poisson, and Poisson-Gaussian noise. For salt-and-pepper noise, we set $\bm{r}\!=\!1$ since the range of the noise is larger.

\vspace{0.2cm}
\noindent\textbf{Obtaining $\bbE \qut{ f_l(\bmz) \mid  \bmyh }$.}
This can be learned in a supervised way by using samples of the pair $(\bmyh, f_l(\bmz))$. To do this, we train a deep neural network $E$ by minimising the mean squared distance between $E(\bmyh)$ and $f_l(\bmz)$. In our implementation, the functions are selected as $f_l(\bmz) = t_l \bmz^l$ for $l = 1, 2, 3$. Here $t_l$ is a scaling constant that is chosen such that $\bbE_{\bmyh} (\frac{1}{n}\left\|\bbE \qut{ f_l(\bmz) \mid  \bmyh } \right\|_2^2)$ is approximately $1$, to prevent the instabilities of training $E$ caused by very small or large magnitudes of network outputs.

\vspace{0.2cm}
\noindent\textbf{Computing $D\qut{\bmyh}$.}
The initial denoising outcomes $D\qut{\bmyh}$ are computed with the given denoising scheme. The denoising scheme doesn't need to produce very accurate results but they will be improved in the refinement process. For Poisson noise, we consider linear filters \cite{gonzalez2008digital} (with a constant convolution kernel of size $5\times 5$), unsupervised learning method Noise2Self \cite{batson2019noise2self}, and the well-established BM3D method \cite{dabov2007image}. For Noise2Self, we use the same network architecture of $R_\theta$ as above. For salt-and-pepper noise, we consider the median filters \cite{gonzalez2008digital}, and the iterative mean filters \cite{thanh2019iterative}. 

\vspace{0.2cm}
\noindent\textbf{Optimisation process and inference.} 
The networks $R_\theta$ and $G_\omega$ are trained by minimising the loss function in \eqref{eq:L2-2} and the optimisation method described in Subsection \ref{subsect:refinement}, where $\gamma$ is increased during the training processing (the specification is given in the supplemental material). In all cases, distance measure $\mathcal{D}$ is specified as the $L_1$ norm of the difference between the two arguments. The updates of $\theta$ and $\gamma$ are implemented using the Adam method \cite{kingma2014adam} with a batch size of $128$. The images are cropped to the size of $40 \times 40$ during training. Each training process consists of 80 epochs, where each epoch contains 1860 optimisation steps. We apply a step decay learning rate scheme, starting from an initial value of $10^{-3}$ in the first 40 epochs, followed by $10^{-4}$ in the next 20 epochs, and $5\times 10^{-5}$ for the rest of the training. \added{In the inference phase, once the networks are trained, the denoised output are obtained by applying $R_\theta$ to the original noisy images $\bmy$ (instead of $\hat{\bmy}$) and averaging the resulting $K$ images. We do not compute $\bmyh$ or the auxiliary vector $\bmz$ during inference.}

\subsection{Improvement of noise consistency}
We start by investigating the statistical noise consistency of existing denoising schemes and their improved versions computed by the proposed refinement method. In this experiment, we focus on Poisson noise with parameter $\lambda=15$ and compare both sides of \eqref{eq:fz2} where $p_{\theta, \bmyh, i}$ is replaced by the respective existing denoising results and $G_\omega(\cdot, \bmyh_i)$ is optimised by minimising $\mathcal{L}_2$. 

\begin{figure*}[ht!]
    \centering
    \setlength{\tabcolsep}{2pt}
    \begin{tabular}{c@{~}@{~}@{~}ccccc} 
      \includegraphics[width=0.135\linewidth,trim={5 5 5 5}, clip]{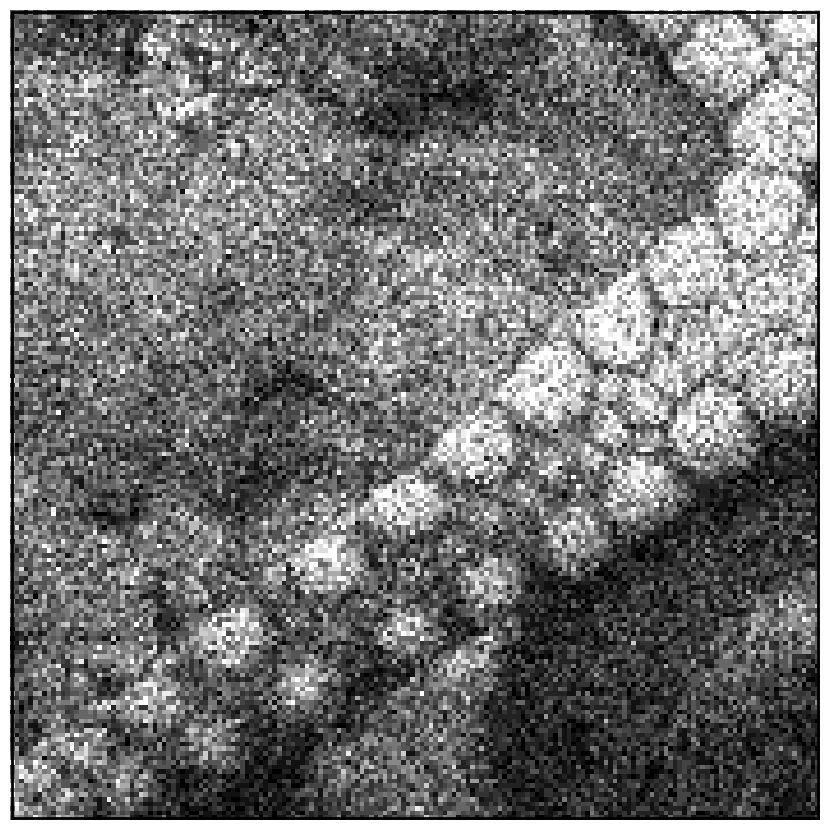}
      & \includegraphics[width=0.16\linewidth,trim={5 5 5 5}, clip]{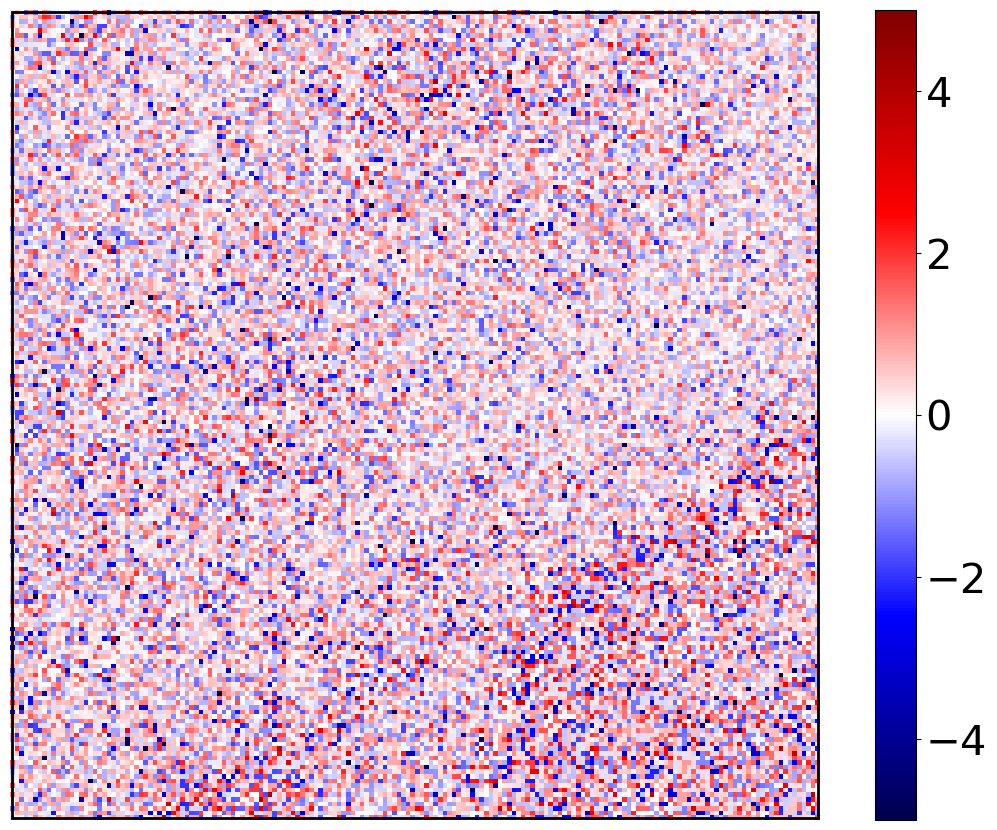}
      & \includegraphics[width=0.16\linewidth,trim={5 5 5 5}, clip]{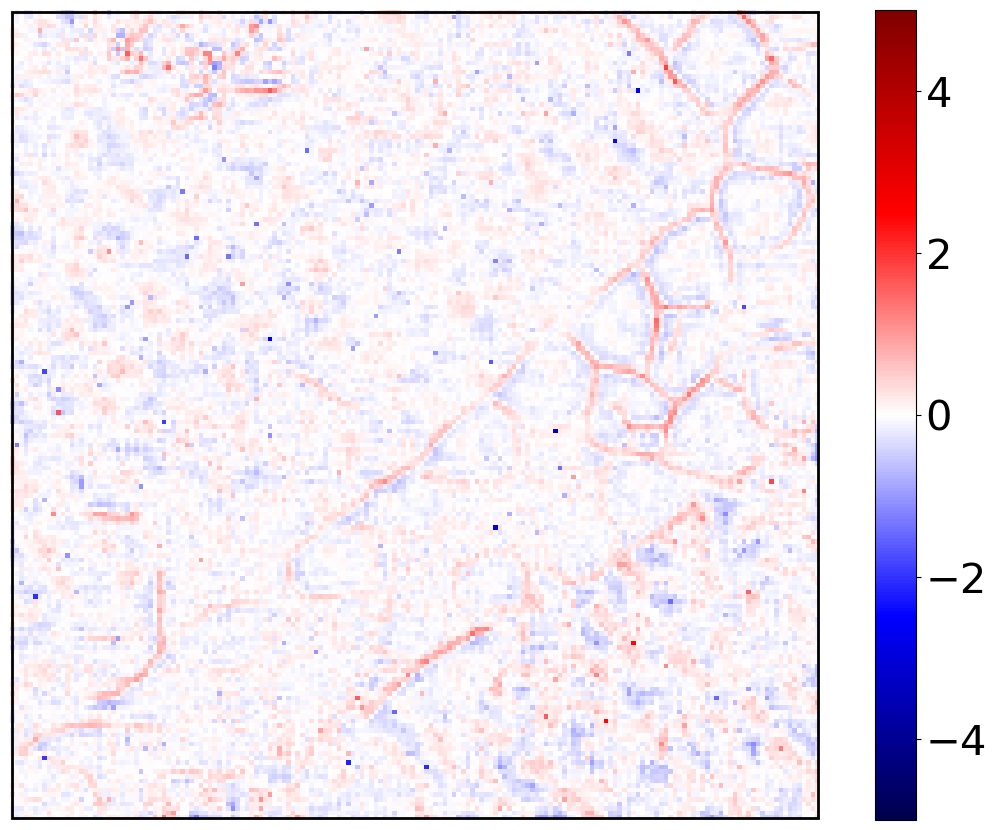}
      & \includegraphics[width=0.16\linewidth,trim={5 5 5 5}, clip]{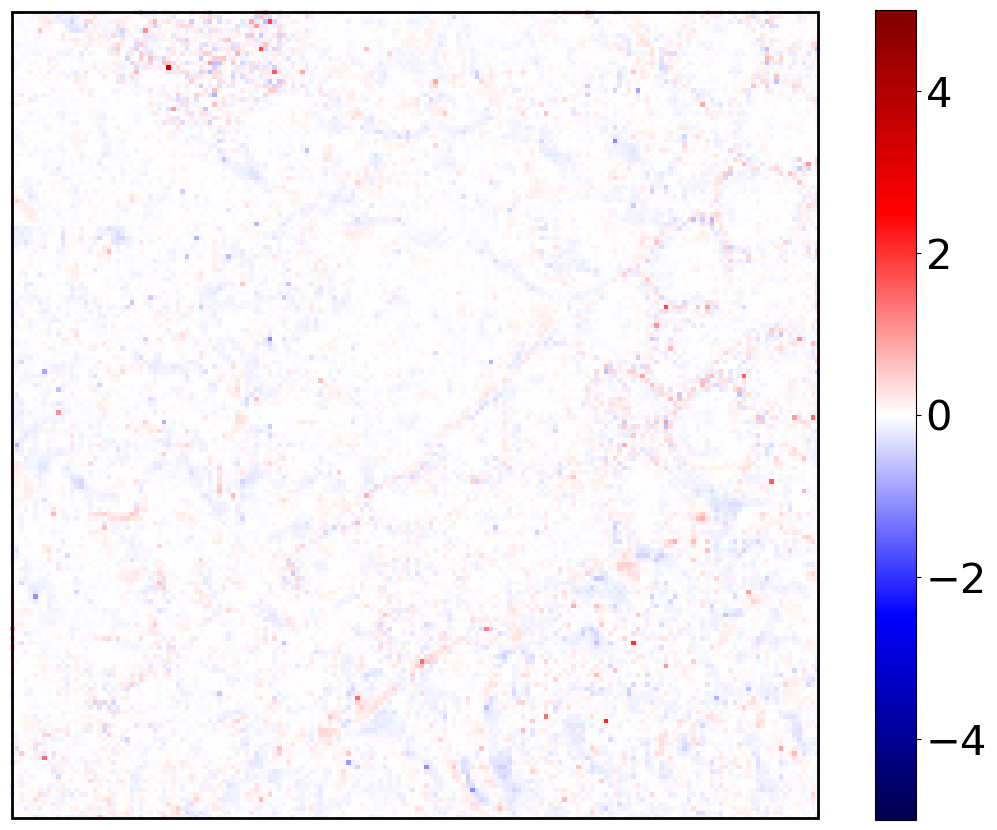}
      & \includegraphics[width=0.16\linewidth,trim={5 5 5 5}, clip]{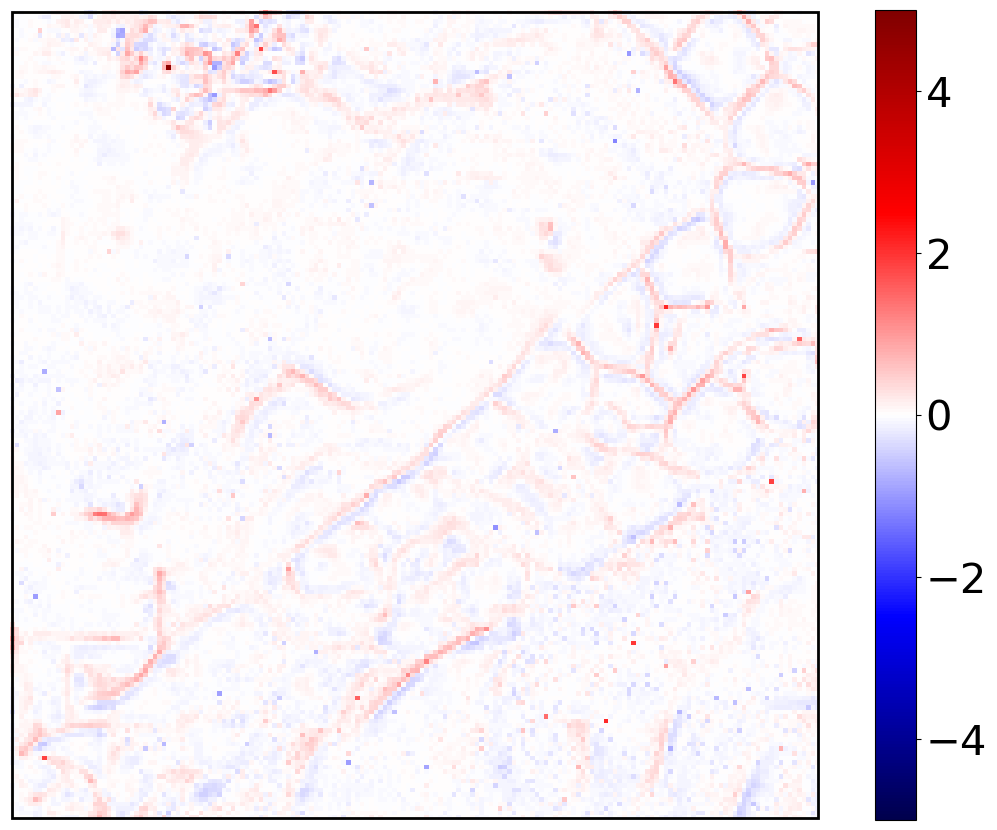}
      & \includegraphics[width=0.16\linewidth,trim={5 5 5 5}, clip]{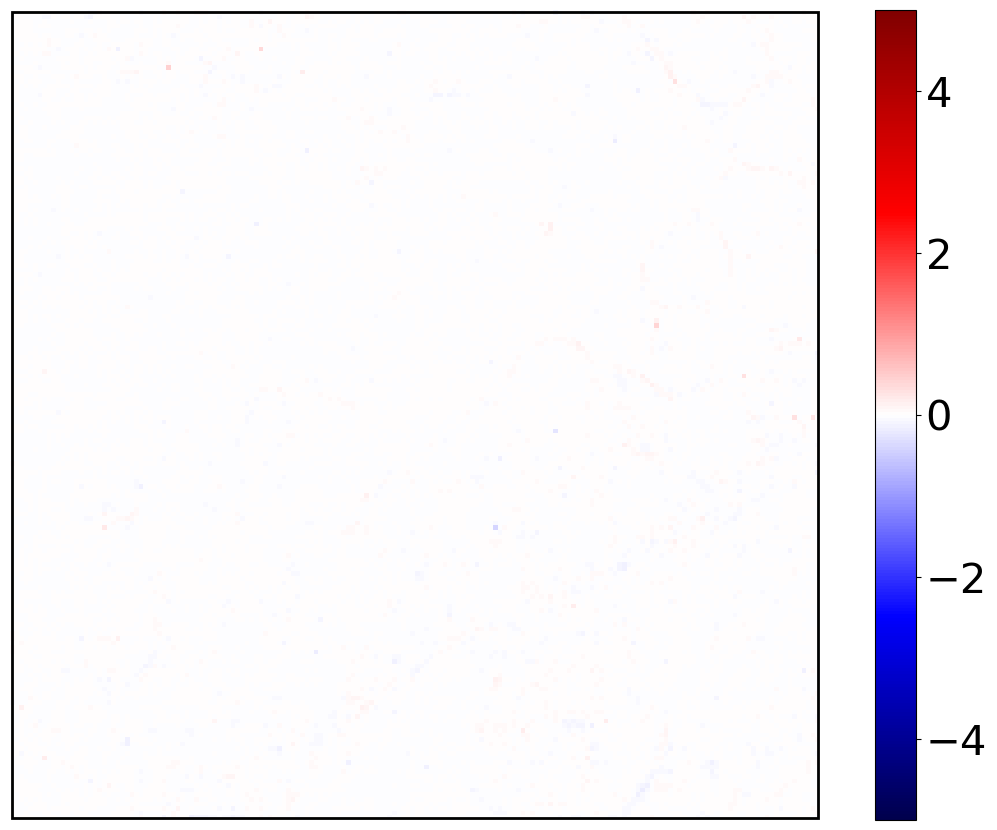}
      \\ 
      \includegraphics[width=0.135\linewidth,trim={5 5 5 5}, clip]{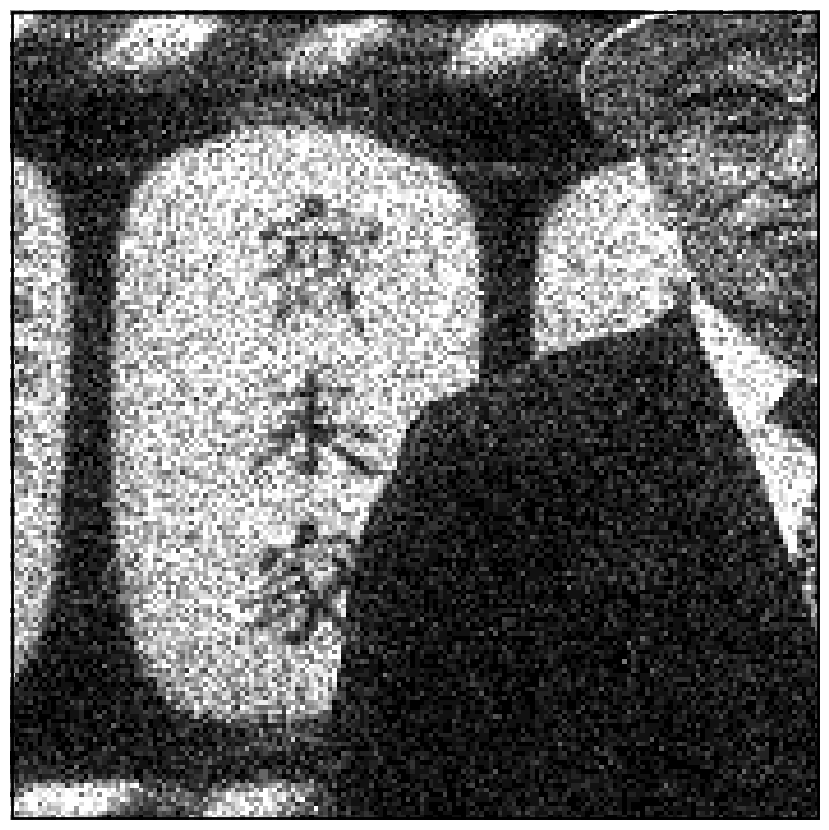}
      & \includegraphics[width=0.16\linewidth,trim={5 5 5 5}, clip]{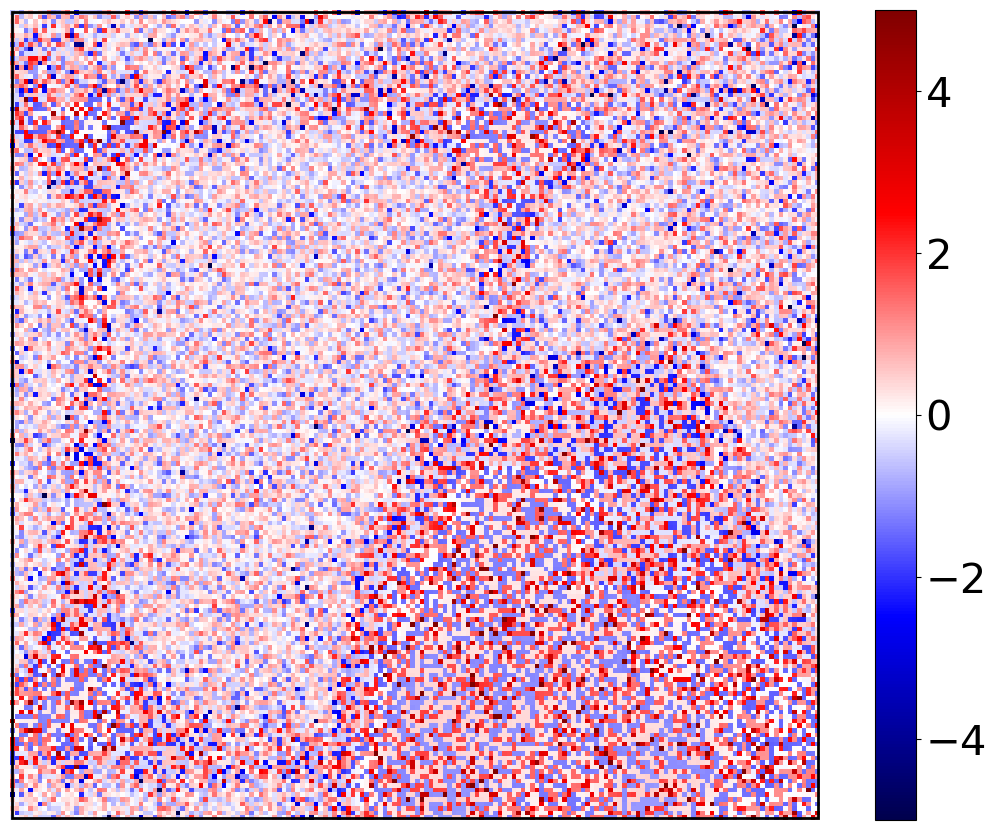}
      & \includegraphics[width=0.16\linewidth,trim={5 5 5 5}, clip]{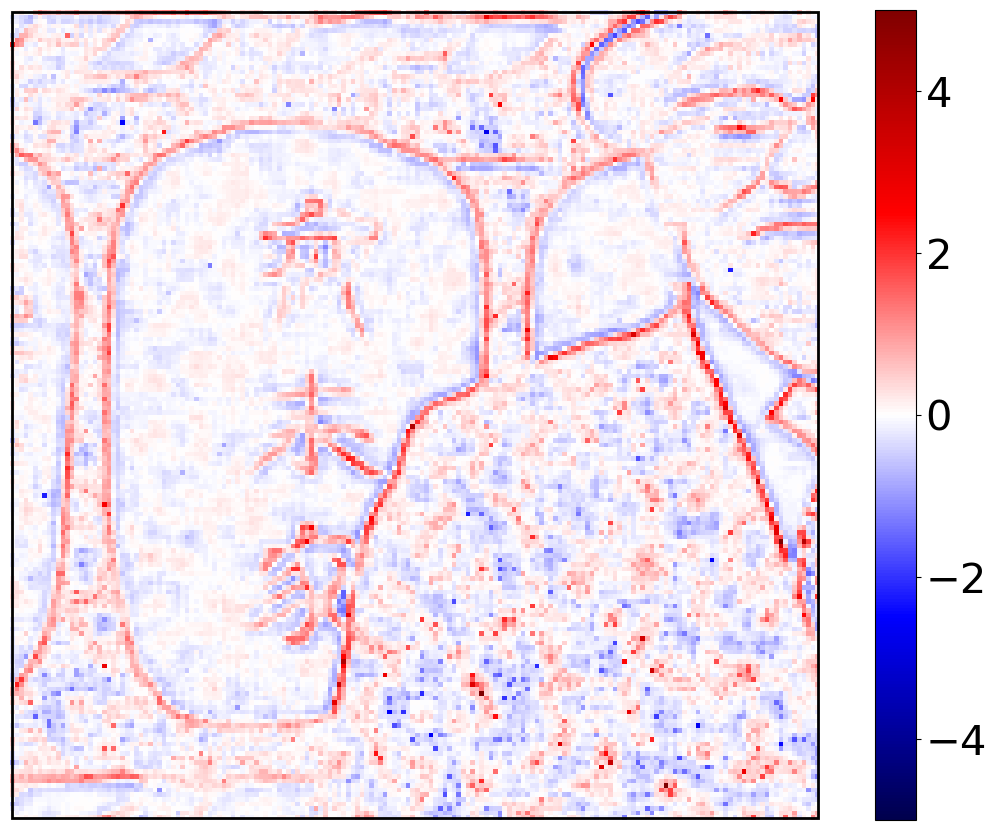}
      & \includegraphics[width=0.16\linewidth,trim={5 5 5 5}, clip]{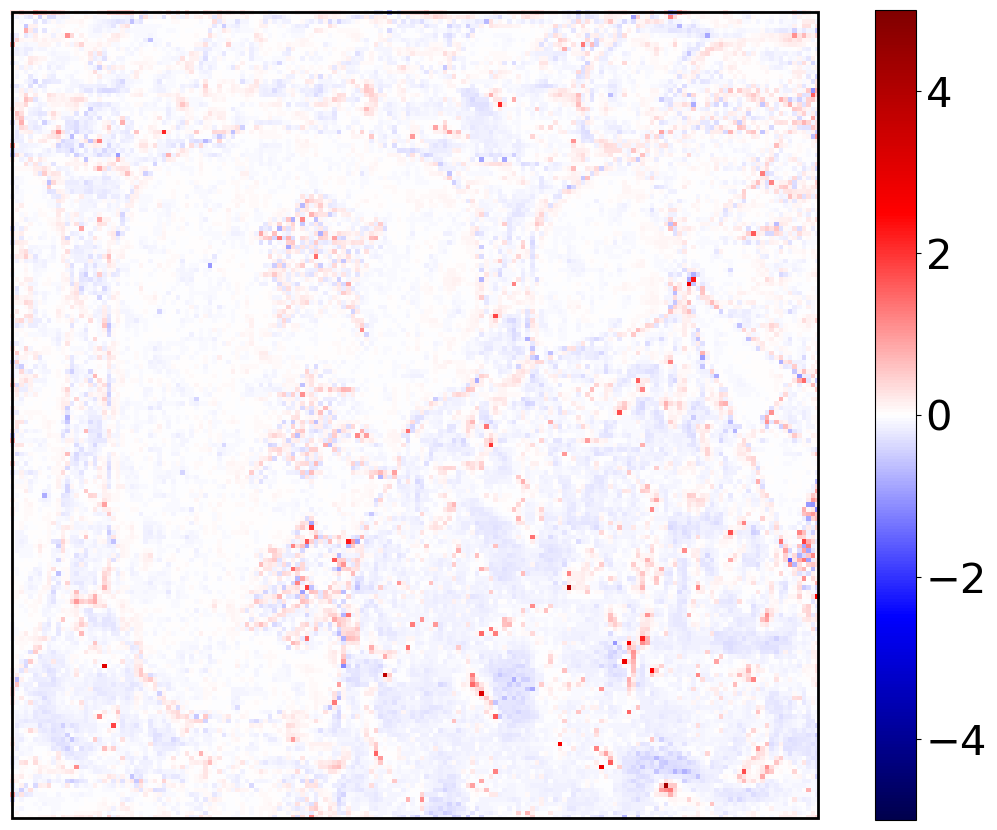}
      & \includegraphics[width=0.16\linewidth,trim={5 5 5 5}, clip]{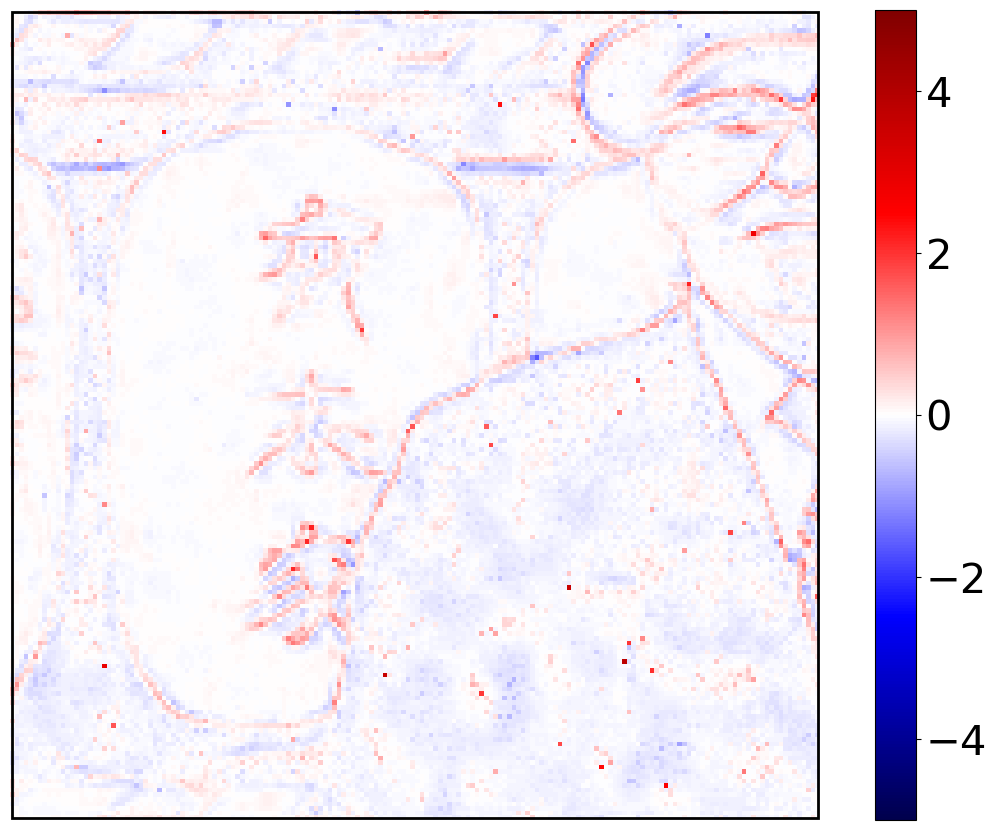}
      & \includegraphics[width=0.16\linewidth,trim={5 5 5 5}, clip]{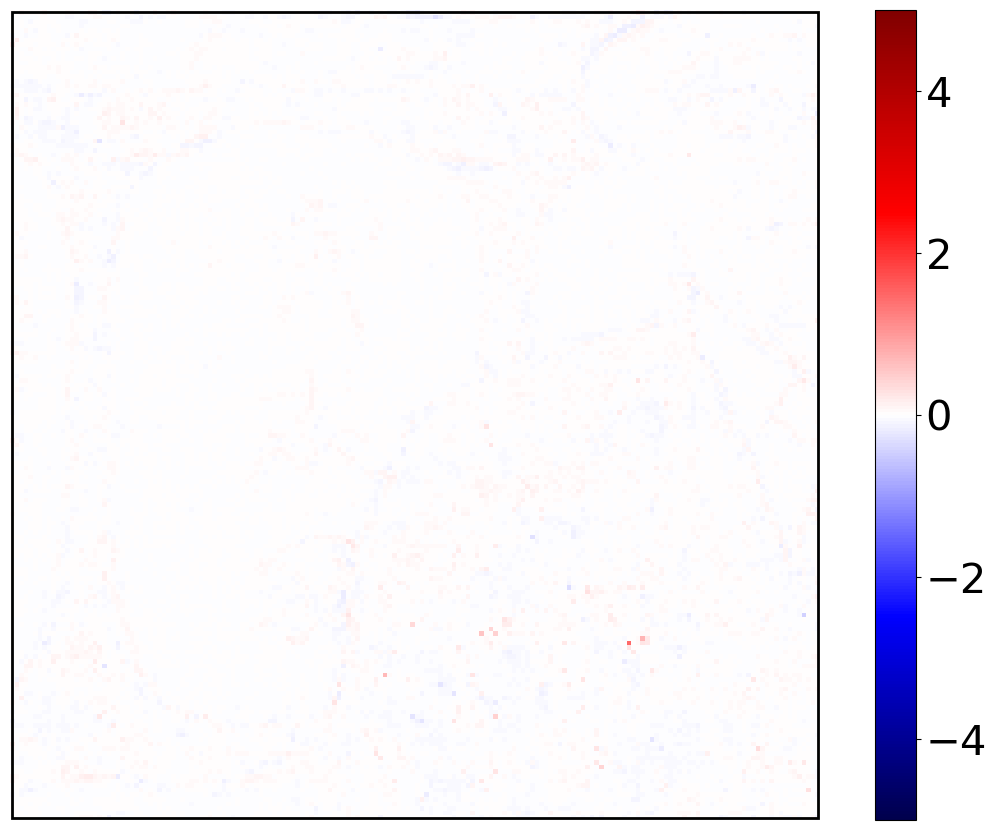}
      \\         
      \includegraphics[width=0.135\linewidth,trim={5 5 5 5}, clip]{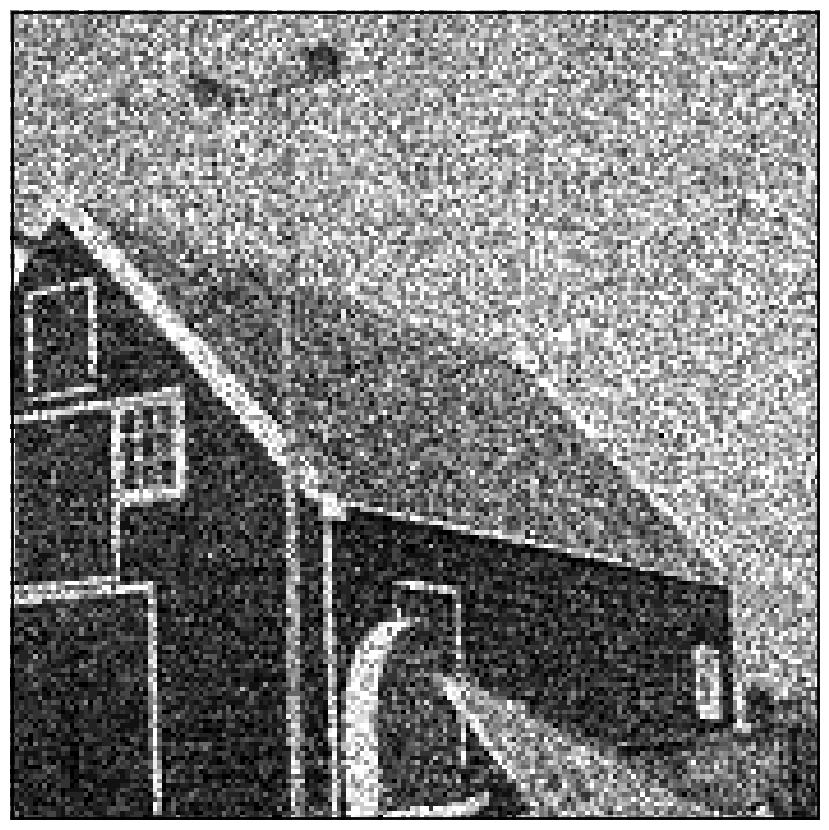}
      & \includegraphics[width=0.16\linewidth,trim={5 5 5 5}, clip]{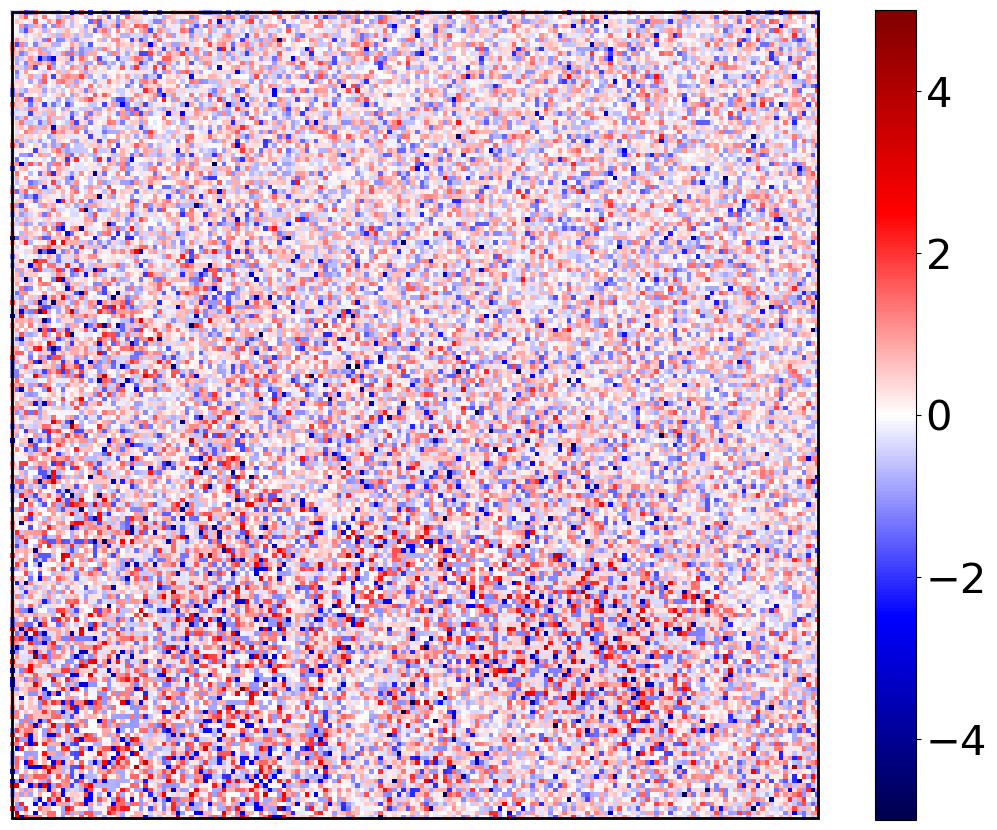}
      & \includegraphics[width=0.16\linewidth,trim={5 5 5 5}, clip]{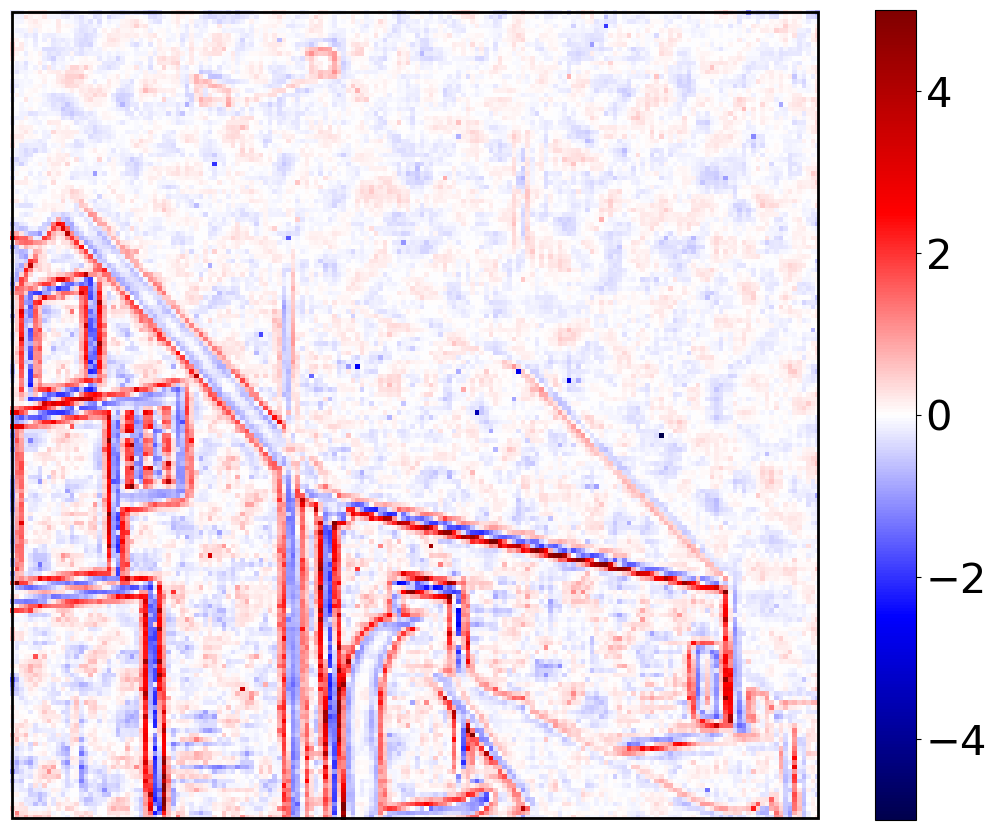}
      & \includegraphics[width=0.16\linewidth,trim={5 5 5 5}, clip]{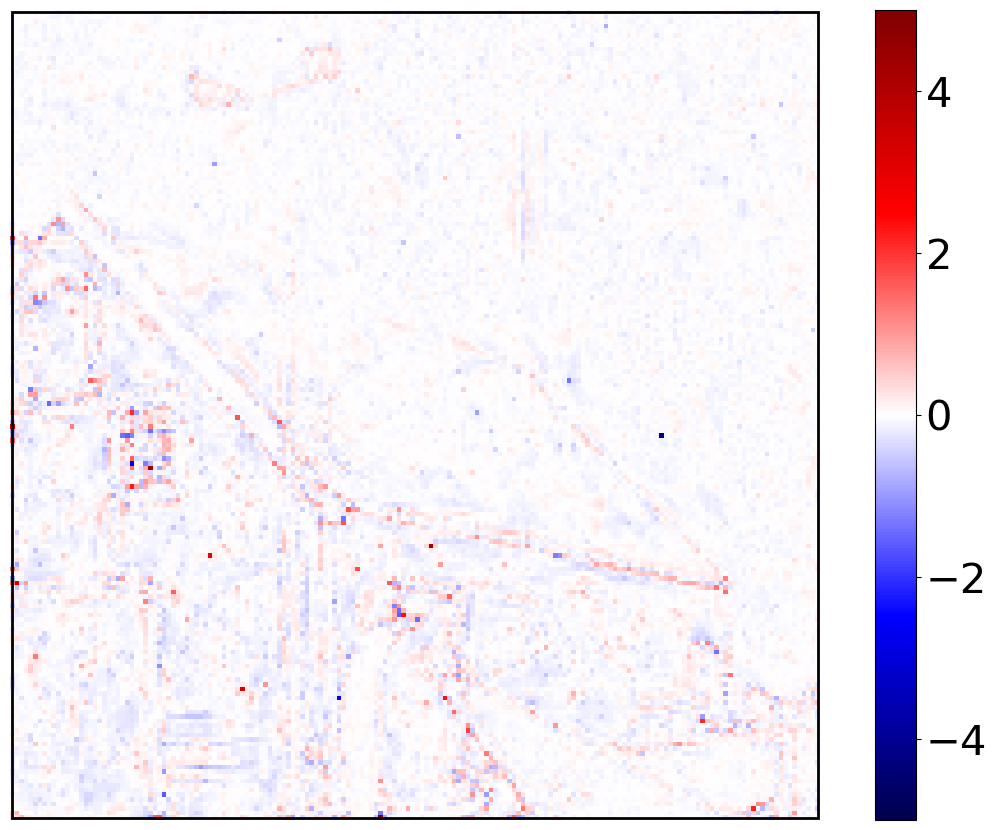}
      & \includegraphics[width=0.16\linewidth,trim={5 5 5 5}, clip]{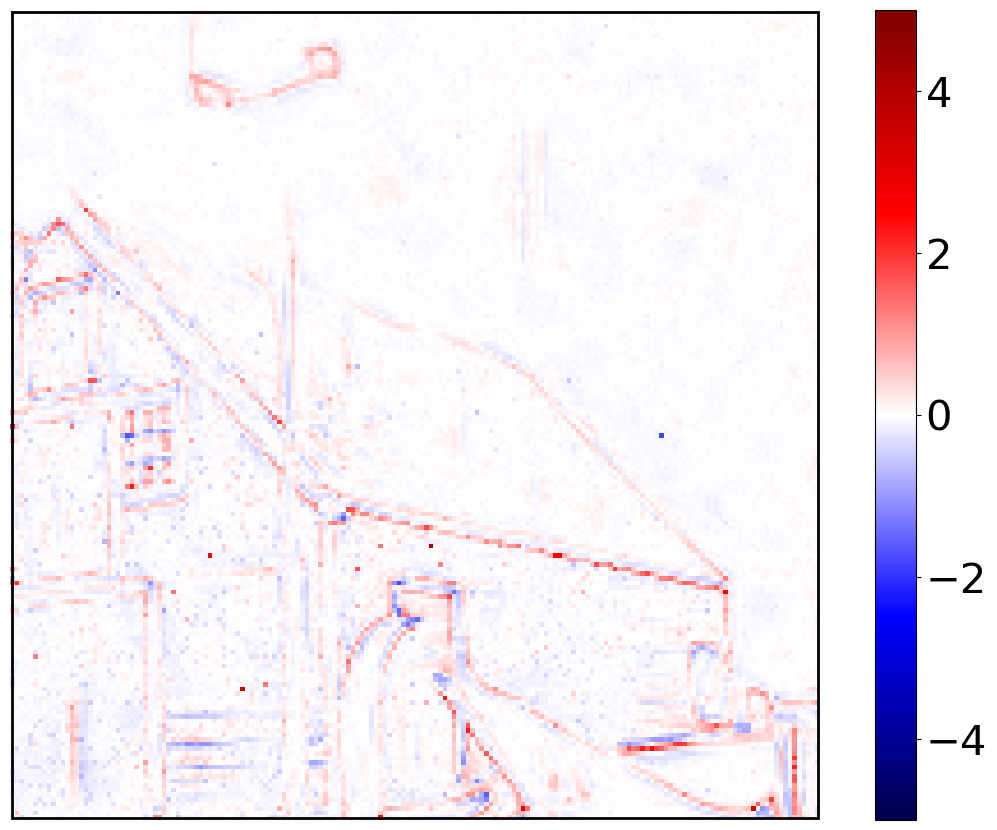}
      & \includegraphics[width=0.16\linewidth,trim={5 5 5 5}, clip]{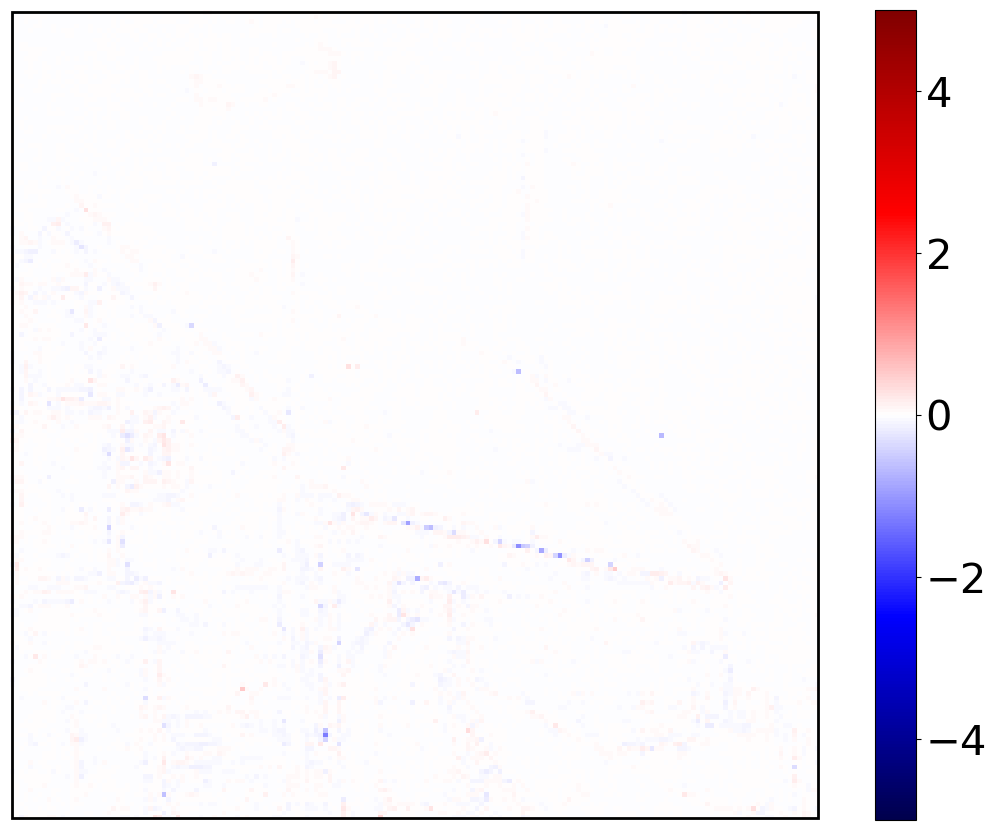}
      \\   
      noisy data $\bmyh$ & $\bbE\qut{f_1(\bmz) \mid \bmyh}$ & LF Residual & N2S Residual & BM3D Residual &  Refined Residual \\
    \end{tabular}
    \caption{Visualisation of the levels of inconsistencies for different schemes. The residuals on the last 4 columns are the difference between $\bbE\qut{f_1(\bmz) \mid \bmyh}$ and the associated predicted expectation using the denoising results of the respective schemes, indicating the inconsistencies with respect to the noise statistics.    
    }    \label{fig:consistency1}
\end{figure*}

\begin{table}[ht!]
  \caption{Denoising quality for Gaussian noise, measured in PSNR (dB)}
  \label{tab:Gau}
  \centering
  \makebox[\linewidth]{
  \begin{tabular}{c||cc|cc}
    \toprule
    data sets & \multicolumn{2}{c||}{BSD68} & \multicolumn{2}{c}{Set12} \\
    \midrule
    \backslashbox[2.5cm]{Methods}{ $\sigma$}
& {$25$} & $50$  & {$25$} & $50$         \\
    \midrule
    \midrule
    LF (Linear filters) \cite{gonzalez2008digital} & 24.49 & 23.08 & 24.62 & 23.29   \\
    Noise2Self \cite{batson2019noise2self} & 27.06 &  25.14 & 28.37 & 26.01  \\    
    BM3D \cite{dabov2007image} & 28.62 & 25.69 & 30.00 & 26.76 \\
    LGSR \cite{zha2022low} & 28.79 & 25.84 & 30.22 & \textbf{27.04}     \\
    Neighbour2Neighbour \cite{huang2022neighbor2neighbor} & 28.87 & 25.85 & 30.02 & 26.74 \\
    MMES \cite{yokota2020manifold} & 27.21 & 24.62 & 28.17 & 25.07  \\
    Noise2Fast \cite{lequyer2022fast} & 27.87 & 24.55 & 29.00 & 25.49   \\
    Refined BM3d & \textbf{29.09} & \textbf{26.11} & \textbf{30.29} & 27.03 \\
    \midrule
    \midrule
    Supervised method & 29.20 & 26.24 & 30.40 & 27.18  \\    
    \bottomrule
  \end{tabular}
  }
\end{table}

\begin{table*}[ht!]
  \caption{Quality of refined denoising results for Poisson noise with parameter $\lambda$ on BSD68 and Set12. The numbers in the parentheses indicate the improvements of the refined models upon their corresponding original versions, measured in PSNR (dB).}
  \label{tab:Poi-psnr}
  \centering
  \makebox[\linewidth]{
  \begin{tabular}{c|c|c|c||c|c|c}
    \toprule
    data sets & \multicolumn{3}{c||}{BSD68} & \multicolumn{3}{c}{Set12} \\
    \midrule
    \backslashbox[2.5cm]{Methods}{ $\lambda$}
& {$15$} & $30$ & $60$ & {$15$} & $30$ & $60$        \\
    \midrule
    \midrule
    LF (Linear Filters) \cite{gonzalez2008digital} & 23.49 & 24.20 & 24.64 & 23.49 & 24.25 & 24.70 \\
    Noise2Self \cite{batson2019noise2self} & 25.49 &  26.49 & 27.64 & 26.37 & 27.63 & 28.98 \\    
    BM3D \cite{dabov2007image} & 25.58 & 26.89 & 28.34 & 26.78 & 28.39 & 29.96\\
    LGSR \cite{zha2022low} & 25.05 & 26.90 & 28.52 & 26.51 & 28.38 & 30.10  \\
    Neighbor2Neighbor \cite{huang2022neighbor2neighbor} & 26.52 & 27.82 & 29.45 & 27.29 & 28.91 & 30.57\\
    MMES \cite{yokota2020manifold} & 24.98 & 26.23 & 27.65 & 25.31 & 26.94 & 28.50 \\
    Noise2Fast \cite{lequyer2022fast} & 25.19 & 26.85 & 28.44 & 26.14 & 27.90 & 29.45 \\
    \midrule
    Refined LF & 26.51 {\footnotesize (+3.02)} & 27.72 {\footnotesize (+3.52)} & 29.16 {\footnotesize (+4.52)} & 27.39 {\footnotesize (+3.90)} & 28.68 {\footnotesize (+4.43)} & 30.10 {\footnotesize (+5.40)} \\
    Refined Noise2Self & 26.19 {\footnotesize(+0.70)} & 27.51 {\footnotesize(+1.02)} & 28.90 {\footnotesize(+1.26)} & 27.10 {\footnotesize(+0.73)} & 28.46 {\footnotesize(+0.83)} & 29.91 {\footnotesize(+0.93)} \\
    Refined BM3D & 26.57 {\footnotesize(+0.99)} & 28.03 {\footnotesize(+1.14)} & 29.50 {\footnotesize(+1.16)} & 27.51 {\footnotesize(+0.73)} & 29.06 {\footnotesize(+0.67)} & 30.61 {\footnotesize(+0.65)} \\
    \added{Refined LGSR} & \added{26.41 {\footnotesize(+1.36)}} & \added{27.86 {\footnotesize(+0.96)}} &  \added{29.23 {\footnotesize (+0.71)}} & \added{27.31 {\footnotesize(+0.80)}} & \added{28.82 {\footnotesize(+0.44)}} &  \added{30.52 {\footnotesize(+0.42)}}\\
    \added{{\footnotesize Refined Neighbor2Neighbor}} & \added{\textbf{26.69} {\footnotesize(+0.17)}} & \added{\textbf{28.21}{\footnotesize(+0.39)}} & \added{\textbf{29.79} {\footnotesize(+0.34)}} & \added{\textbf{27.56} {\footnotesize(+0.27)}} & \added{\textbf{29.17} {\footnotesize(+0.26)}} & \added{\textbf{30.80} {\footnotesize(+0.23)}} \\    
    \added{Refined MMSE} & \added{26.21 {\footnotesize(+1.22)}} & \added{27.81 {\footnotesize(+1.58)}} & \added{29.26 {\footnotesize(+1.61)}} & \added{27.22 {\footnotesize(+1.91)}} & \added{28.67 {\footnotesize(+1.73)}} & \added{30.14 {\footnotesize(+1.64)}}\\
    \added{Refined Noise2Fast} & \added{26.26 {\footnotesize(+1.07)}} & \added{27.92 {\footnotesize(+1.07)}} & \added{29.60 {\footnotesize(+1.16)}} & \added{27.39 {\footnotesize(+1.25)}} & \added{29.06 {\footnotesize(+1.16)}} & \added{30.71 {\footnotesize(+1.26)}} \\
    
    \midrule
    \midrule
    Supervised baseline & 26.84 & 28.37 & 29.96 & 27.73 & 29.32 & 30.85 \\    
    \bottomrule
  \end{tabular}
  }
\end{table*}

Figure \ref{fig:consistency1} shows the results for two images taken from the training set. It visualises the level of inconsistencies measured by the residual $\bbE \qut{ f(\bmz_i) \mid  \bmyh } - \langle p_{\theta, \bmyh, i}(\cdot),  G_\omega(\cdot, \bmyh_i) \rangle$. As illustrated in Figure \ref{fig:consistency1}, the linear filter (LF) produces the largest level of inconsistencies, particularly near the edge structures. The Noise2Self (N2S) \cite{batson2019noise2self} and BM3D have better consistency, but there are still image structures presented in the residuals (in columns 4 and 5). The residuals of the refined version for LF (in the last column) are much smaller than the others, hence the refined version is more consistent with the noise statistics. The denoising qualities will be compared next.

\subsection{Refinement results and comparison}
We will compare the denoising qualities before and after the refinement process and report the performance gap between the obtained results and supervised learning baselines. The quality of the denoised images is measured using the average scores of the peak signal-to-noise ratio (PSNR) and structural similarity index measure (SSIM) \cite{wang2004image} across the test sets (BSD68 \cite{martin2001database} and Set12 \cite{zhang2017beyond}). We report the results for Poisson noise and salt-and-pepper noise respectively. 

\vspace{0.2cm}
\noindent\textbf{Gaussian Noise.} The results for Gaussian noise are summerised in Table \ref{tab:Gau}, where our method is denoted as Refined BM3D (a refined version of BM3D) and compared against unsupervised methods including Noise2Self \cite{batson2019noise2self}, LGSR \cite{zha2022low}, Neighbor2Neighbor \cite{huang2022neighbor2neighbor}, MMES \cite{yokota2020manifold}, and Noise2Fast \cite{lequyer2022fast}. It is shown that Refined BM3D significantly improves upon BM3D and outperforms the unsupervised methods for different $\sigma$ values (which denotes the standard deviation of Gaussian noise, relative to 256 intensity levels).  Additionally, the performance of Refined BM3D is close to that of the supervised model (based on exactly the same network architecture), which shows that the quality of our results is similar to their supervised counterparts obtained with ground truth training data.

\vspace{0.2cm}
\noindent\textbf{Poisson Noise.} The results for Poisson noise are compared in Table \ref{tab:Poi-psnr} for different noise parameters $\lambda$$=$$15,30,60$. The results illustrate that the refinement process significantly improves the quality of the denoising results for all cases. The LF (Linear Filter) method yields low-quality results, but these are improved by more than $3$ dB by the proposed method, outperforming other denoisers like BM3D and Noise2Self. The substantial improvement is due to the fact that LF is a naive approach and its results exhibit a high level of inconsistencies (also see Figure \ref{fig:consistency1}). As expected, after the refinement process, the performance gap between LF and BM3D becomes smaller due to the higher level of consistency. 

Although BM3D is not designed to remove Poisson noise, this disadvantage is mitigated by the refinement approach, which consistently leads to a 0.6 dB to 1.2 dB improvement across the different cases. Notably, the PSNR values of refined BM3D are only about $0.3$ dB to $0.5$ dB less than that of the supervised method (which produces the optimal results as it uses ground truth data for training) and outperforms Neighbor2Neighbor \cite{huang2022neighbor2neighbor} (a customised unsupervised approach for zero-mean noise). The small performance gap suggests that, by imposing the consistency constraint in \eqref{eq:unconstrainedopt}, the Refined BM3D is able to capture further information of the clean data that was lost in the original BM3D. \added{When applied to refining Neighbor2Neighbor, our method further reduces the performance gap to the supervised baseline to around $0.2$ dB, demonstrating its effectiveness as a refinement approach.}

\begin{figure*}[ht!]
    \centering
    \small 
    \setlength{\tabcolsep}{2pt}
    \begin{tabular}{cccccc} 
      \includegraphics[width=0.155\linewidth,trim={5 5 5 5}, clip]{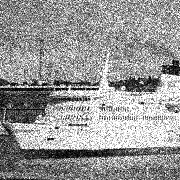}
      &
      \includegraphics[width=0.155\linewidth,trim={5 5 5 5}, clip]{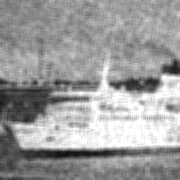} &
      \includegraphics[width=0.155\linewidth,trim={5 5 5 5}, clip]{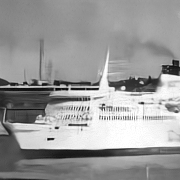} &
      \includegraphics[width=0.155\linewidth,trim={5 5 5 5}, clip]{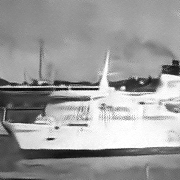}&
      \includegraphics[width=0.155\linewidth,trim={5 5 5 5}, clip]{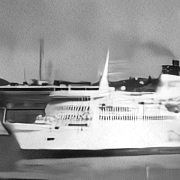} &
      \includegraphics[width=0.155\linewidth,trim={5 5 5 5}, clip]{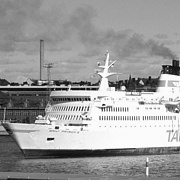} 
      \\ 
      \includegraphics[width=0.155\linewidth,trim={5 5 5 5}, clip]{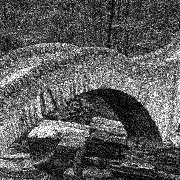}
      &
      \includegraphics[width=0.155\linewidth,trim={5 5 5 5}, clip]{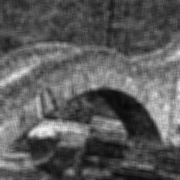} &
      \includegraphics[width=0.155\linewidth,trim={5 5 5 5}, clip]{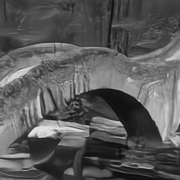} &
      \includegraphics[width=0.155\linewidth,trim={5 5 5 5}, clip]{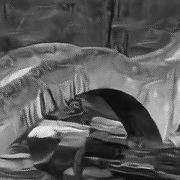}&
      \includegraphics[width=0.155\linewidth,trim={5 5 5 5}, clip]{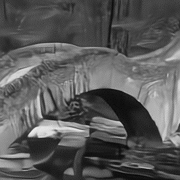} &
      \includegraphics[width=0.155\linewidth,trim={5 5 5 5}, clip]{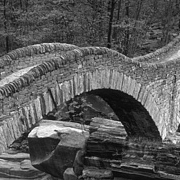} 
      \\ 
      \includegraphics[width=0.155\linewidth,trim={5 5 5 5}, clip]{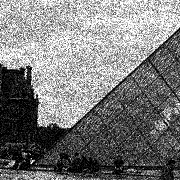} &
      \includegraphics[width=0.155\linewidth,trim={5 5 5 5}, clip]{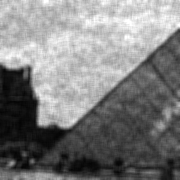} &
      \includegraphics[width=0.155\linewidth,trim={5 5 5 5}, clip]{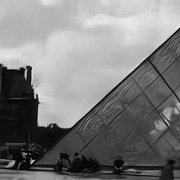} &
      \includegraphics[width=0.155\linewidth,trim={5 5 5 5}, clip]{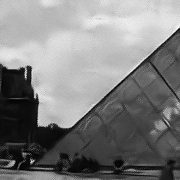}&
      \includegraphics[width=0.155\linewidth,trim={5 5 5 5}, clip]{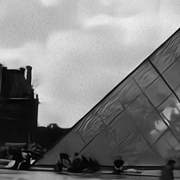} &
      \includegraphics[width=0.155\linewidth,trim={5 5 5 5}, clip]{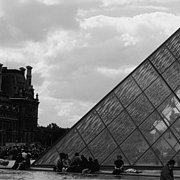} 
      \\   
      noisy data & LF & LF-Refined & N2S & N2S-Refined & Ground truth \\
    \end{tabular}
    \caption{Visualisation of the refined results for examples taken from the training dataset (Poisson noise with $\lambda = 15$). The LF- and N2S-Refined mean the refined results. (Best viewed with Zooming) } \label{fig:quality-Poission}
\end{figure*}

Figure \ref{fig:quality-Poission} compares the qualitative results of the difference schemes, focusing on Poisson noise with $\lambda = 15$.  The three examples are taken from the training set to illustrate the results of the refinement process during the training. According to the figure, although the resulting images of Linear Filter (2nd column) look blurry and noisy, their refined version recovers the fine details well. The N2S results recover well the smooth region of the images, but some details of edges and thin objects are not visible (for example the windows of the boat on the first row). The refined version NS2-Refined (column 6) shows better results on these details.

\vspace{0.2cm}
\noindent\textbf{Salt-and-pepper noise.} The test results for different levels of salt-and-pepper noise are reported in Table \ref{tab:SP-psnr}, where the noise level corresponds to the proportion of corrupted pixels in the noisy data. We observe significant improvements in PSNR after the refinement for the basic median filter (MF) \cite{gonzalez2008digital}, by around 10 dB for $10\%$ and $30\%$ noise, and by around 7 dB for $50\%$ noise. Interestingly, the quality of the refined results outperform the sophisticated method IMF \cite{thanh2019iterative} for removing salt-and-pepper noise by 2dB to 5dB. The better results are achieved by the refined IMF, leading to a small performance gap (from 0.3dB to 1dB across the different noise level) compared with the supervised method which is optimal in the sense that it learns from the clean data. 

\begin{table*}[ht]
  \caption{Quality of refined denoising results for salt-and-pepper noise on BSD68 and Set12. Here ``Refined MF" and ``Refined IMF" denote the refined results with the proposed method. The associated improvements in PSNR (dB) are given in parentheses.}
  \label{tab:SP-psnr}
  \centering
  \makebox[\linewidth]{
  \begin{tabular}{c|c||cc|cc|cc}
    \toprule

& Noise Level 
 & \multicolumn{2}{c|}{$10$\%} & \multicolumn{2}{c|}{$30$\%} & \multicolumn{2}{c}{$50$\%}  \\
& Measure & PSNR & SSIM & PSNR & SSIM & PSNR & SSIM \\
\midrule
\midrule
 \multirow{5}{1em}{\rotatebox[origin=c]{90}{Dataset: BSD68}}   
    & Median Filters (MF) & 28.29 & 0.8975 & 24.93 & 0.7986 & 23.20 & 0.7348  \\
    & IMF \cite{thanh2019iterative} & 36.97 & 0.9889 & 31.76 & 0.9625 & 28.97 & 0.9270 \\
    & Refined MF & 41.10 (+12.81) & \textbf{0.9945} & 33.90 (+8.97) & 0.9735 & 30.59 (+7.39)  & 0.9360 \\
    & Refined IMF & \textbf{41.38} (+4.41) & {0.9944} & \textbf{35.19} (+3.43) & \textbf{0.9770} & \textbf{31.41} (+2.44) & \textbf{0.9472} \\
    \cmidrule{2-8}
    & Supervised baseline & 42.26 & 0.9953 & 35.90 & 0.9806 & 32.28 & 0.9567 \\ 
    \midrule
    \midrule
 \multirow{5}{1em}{\rotatebox[origin=c]{90}{Dataset: Set12}}   
    & Median Filters (MF) & 29.12 & 0.9321 & 24.94 & 0.8592 & 22.89 & 0.8009 \\
    & IMF \cite{thanh2019iterative} & 36.45 & 0.9915 & 31.61 & 0.9723 & 28.93 & 0.9470 \\
    & Refined MF & 41.60 (+12.48) & \textbf{0.9955} & 34.78 (+9.84) & 0.9832 & 31.61 (+8.72)  & 0.9539 \\    
    & Refined IMF & \textbf{41.66} (+5.21)   & \textbf{0.9955} & \textbf{36.04} (+4.43) & \textbf{0.9847} & \textbf{32.53} (+3.60) & \textbf{0.9673} \\
    \cmidrule{2-8}
    & Supervised baseline & 42.85 & 0.9967 & 36.75 & 0.9876 & 33.55 & 0.9751 \\ 
    \bottomrule
  \end{tabular}
  }
\end{table*}

\begin{figure*}[ht!]
    \centering
    \small 
    \setlength{\tabcolsep}{2pt}
    \begin{tabular}{cccccc} 
      \includegraphics[width=0.165\linewidth,trim={5 5 5 5}, clip]{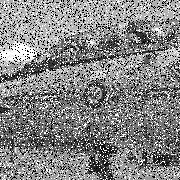}
      &
      \includegraphics[width=0.165\linewidth,trim={5 5 5 5}, clip]{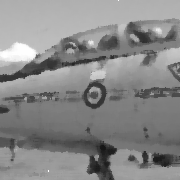} &
      \includegraphics[width=0.165\linewidth,trim={5 5 5 5}, clip]{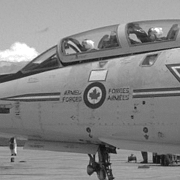} &
      \includegraphics[width=0.165\linewidth,trim={5 5 5 5}, clip]{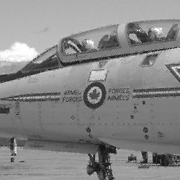}&
      \includegraphics[width=0.165\linewidth,trim={5 5 5 5}, clip]{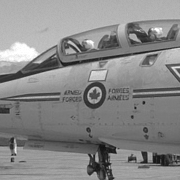} &
      \includegraphics[width=0.165\linewidth,trim={5 5 5 5}, clip]{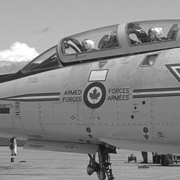} 
      \\ 
      
      \includegraphics[width=0.165\linewidth,trim={5 5 5 5}, clip]{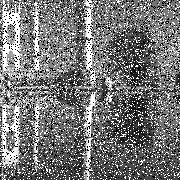}
      &
      \includegraphics[width=0.165\linewidth,trim={5 5 5 5}, clip]{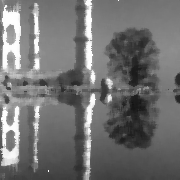} &
      \includegraphics[width=0.165\linewidth,trim={5 5 5 5}, clip]{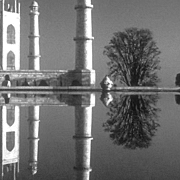} &
      \includegraphics[width=0.165\linewidth,trim={5 5 5 5}, clip]{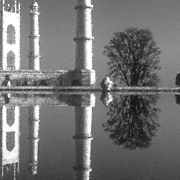}&
      \includegraphics[width=0.165\linewidth,trim={5 5 5 5}, clip]{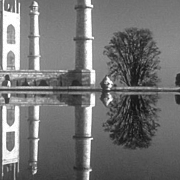} &
      \includegraphics[width=0.165\linewidth,trim={5 5 5 5}, clip]{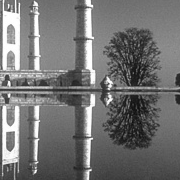} 
    \\
      \includegraphics[width=0.165\linewidth,trim={5 5 5 5}, clip]{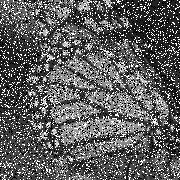}
      &
      \includegraphics[width=0.165\linewidth,trim={5 5 5 5}, clip]{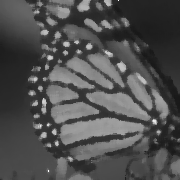} &
      \includegraphics[width=0.165\linewidth,trim={5 5 5 5}, clip]{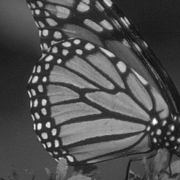} &
      \includegraphics[width=0.165\linewidth,trim={5 5 5 5}, clip]{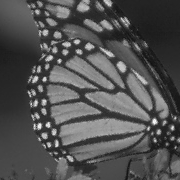}&
      \includegraphics[width=0.165\linewidth,trim={5 5 5 5}, clip]{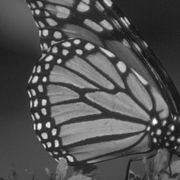} &
      \includegraphics[width=0.165\linewidth,trim={5 5 5 5}, clip]{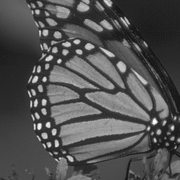} 
      \\   
      noisy data & MF & MF-Refined & IMF & IMF-Refined & Ground truth \\
    \end{tabular}
    \caption{Refinement results for the median filter (MF) and IMF (best viewed with Zooming). The examples are taken from the training data with $30\%$ salt-and-pepper noise. The noisy data is given in the first column and the ground truth images are given in the last column. The 2nd to 5th columns are the denoising results and their refined versions for comparison.}   \label{fig:SP-figure}
\end{figure*}

The denoising results of the median filter (MF) and iterative mean filter (IMF) \cite{thanh2019iterative} are displayed in Figure \ref{fig:SP-figure}. As shown in the figure, the median filter produces results that are low quality and blurry with distorted edges. This also explain the low PSNR values of MF given in Table \ref{tab:SP-psnr}. The MF-refined (refined results of MF) demonstrate a remarkable improvement in the visual quality (for example, the details of the trees in the image on the 2nd row). The denoised images from IMF are not as clear as MF-refined, but their refined versions (IMF-Refined) look visually close to the ground truth image.

\vspace{0.2cm}
\noindent\textbf{Mixed noise.} We consider different levels of Poisson-Gaussian noise, where with $\lambda$ and $\sigma$ being parameters of the Poisson distribution and Gaussian distribution, respectively. \added{As Table \ref{tab:Gau-Poi-psnr} shows, the proposed method significantly improves the quality of existing denoisers. Refined BM3D consistently achieves a small performance gap (around 0.2$-$0.3 dB) compared to its supervised counterparts (with the same denoising network architecture),} which suggests that our method also works for mixed noise. Refined LF (a refined version of the linear filters) performs slightly worse than Refined BM3D, but this is expected because the original results of linear filters are not comparable to those of BM3D. However, we observe that the gap between linear filters and BM3D is much smaller after the refinement, and it is worth noting that, even combined with very simple denoisers, the refined results from our method are significantly better than the other unsupervised approaches for different noise levels.

\begin{table}[ht!]
  \caption{Results (PSNR values) for mixed Poisson–Gaussian noise on data sets BSD68 and Set12 respectively (seperated by $/$). }
  \label{tab:Gau-Poi-psnr}
  \addtolength{\tabcolsep}{-1.5pt}
  \centering
  \makebox[\linewidth]{
  \begin{tabular}{c||ccc}
    \toprule
    $(\lambda, \sigma)$ & $(15, 1.5)$ & $(30,3)$  & $(60, 6)$     \\
    \midrule
    \midrule
    {\footnotesize LF (Linear filters)} \cite{gonzalez2008digital} & 23.07 $/$ 23.09  & 23.69 $/$ 23.77 & 24.05 $/$ 24.17   \\
    BM3D \cite{dabov2007image} & 25.19 $/$ 26.24 & 26.14 $/$ 27.44  & 26.95 $/$ 28.36  \\
    LGSR \cite{zha2022low} &  24.74 $/$ 26.30 &  26.12 $/$ 27.60 & 27.11 $/$  28.61  \\
    Noise2Fast \cite{lequyer2022fast} &  24.43 $/$ 25.30 &  25.62 $/$ 26.58 & 26.47 $/$ 27.56  \\
    Refined LF & 25.93 $/$ 26.79 & 26.86 $/$ 27.87 & 27.48 $/$ 28.60 \\
    Refined BM3D & \textbf{25.95} $/$ \textbf{26.82} & \textbf{26.92} $/$ \textbf{27.93} & \textbf{27.70} $/$ \textbf{28.85} \\
    \added{Refined Noise2Fast} & \added{25.76 $/$ 26.74} & \added{26.81 $/$ 27.88} &  \added{27.65 $/$ 28.83} \\    
    \added{Refined LGSR} & \added{25.86 $/$ 26.73} & \added{26.80 $/$ 27.80} & \added{27.59 $/$ 28.74} \\
    \midrule
    Supervised method & 26.21 $/$ 27.03 & 27.17 $/$ 28.16 & 27.90 $/$ 29.01  \\    
     \bottomrule
  \end{tabular}
  }
\end{table}

\subsection{Performance analysis: computation and hyperparameters}

The proposed method requires two additional networks, $G_\omega$ and $E$, which are used to compute the consistency loss $\mathcal{L}_2$ and to evaluate $\mathbb{E}\left[ f(\bmz) \mid \bmyh \right]$, respectively. %
\added{Such a consistency loss is not required in supervised learning due to the availability of ground-truth images. A comparison of training times for different methods is presented in Table~\ref{tab:train_time}. In our experiment, the proposed method requires approximately $2.5$ times longer training time than the supervised method. However, this additional cost reflects the increased model capacity, which enables our method to learn effectively without access to ground-truth data.}

It should be noted that neither $G_\omega$ nor $E$ is required once the denoising model has been trained. Therefore, there is no extra computational cost during inference. The average computational time over the images in Set12 is reported and compared in Table~\ref{tab:comp_time}. Since our method does not require test-time optimisation, it is significantly faster than BM3D and MMES, and comparable in speed to Neighbor2Neighbor, which uses the same network architecture.

\begin{table}[t]
\centering
\caption{\added{Comparison of training time (in hours)}}
\label{tab:train_time}
\begin{tabular}{l c@{\hskip 1.5em} l c}
\hline
\added{\textbf{Method}} & \added{\textbf{Time (h)}} & \added{\textbf{Method}} & \added{\textbf{Time (h)}} \\
\hline
\added{Supervised baseline \cite{zhang2017beyond}} & \added{6.6} & \added{Noise2Self \cite{batson2019noise2self}} & \added{7.0} \\
\added{Neighbor2Neighbor \cite{huang2022neighbor2neighbor}} & \added{13.1} & \added{The proposed method} & \added{16.1} \\
\hline
\end{tabular}
\end{table}

\begin{table}[t]
\centering
\caption{Average inference time (in seconds) on Set12}
\label{tab:comp_time}
\begin{tabular}{l c@{\hskip 1.5em} l c}
\hline
\textbf{Method} & \textbf{Time (s)} & \textbf{Method} & \textbf{Time (s)} \\
\hline
BM3D \cite{dabov2007image}   & 3.9312  & Noise2Fast \cite{lequyer2022fast}        & 11.92   \\
MMES \cite{yokota2020manifold}              & 291.00  & Neighbor2Neighbor \cite{huang2022neighbor2neighbor} & 0.0199  \\
LGSR \cite{zha2022low} &  1518.95 & Proposed method        & 0.0195 \\
\hline
\end{tabular}
\end{table}

\begin{figure}[ht!]
    \centering
    \small 
    \begin{tabular}{c} 
    \includegraphics[width=0.9\linewidth,trim={0 12 0 0}, clip]{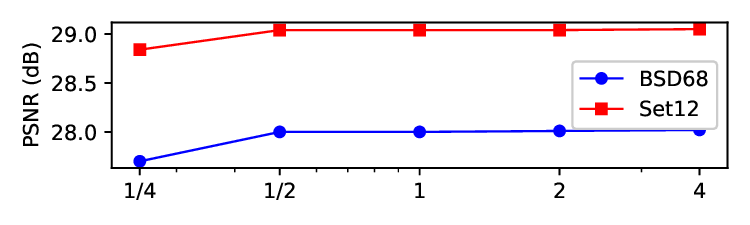} \\
    (a). relative total iteration counts \\
    \includegraphics[width=0.9\linewidth,trim={0 12 0 0}, clip]{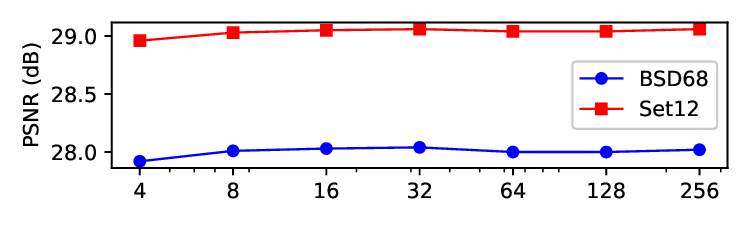} \\
    (b). batch size\\
    \includegraphics[width=0.9\linewidth,trim={0 12 0 0}, clip]{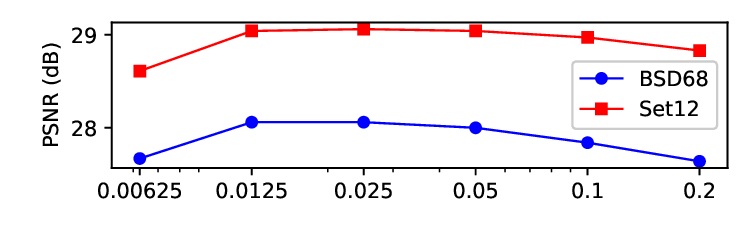} \\
    (c). standard deviation of the variable $\bmz$ \\
    \includegraphics[width=0.9\linewidth,trim={0 12 0 0}, clip]{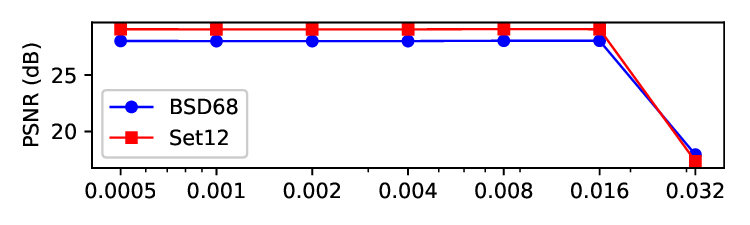} \\
    (d). initial learning rates\\
    \end{tabular}
    \caption{Analysis of performance on test sets over different settings.}   \label{fig:parameters}
\end{figure}

In Figure \ref{fig:parameters}, we study the performance of the model under different hyperparameter settings. The results are obtained under the Poisson noise setting with $\lambda = 30$. As we vary the total number of optimisation steps while keeping other hyperparameters fixed, the denoising quality remains high once the number reaches approximately half of the previously set value (cf. Figure \ref{fig:parameters}(a)). The quality of the results is robust to changes in batch size, as shown in Figure \ref{fig:parameters}(b); however, when the batch size is very small, the training process may take longer due to underutilisation of GPU capacity. Figure \ref{fig:parameters}(c) demonstrates that the PSNR values remain stable within a certain range of the standard deviation for the variable $\bmz$. Performance drops when the standard deviation becomes either too large or too small, which is consistent with our earlier analysis. Finally, similar results are obtained across a range of learning rates (see Figure \ref{fig:parameters}(d)). The model fails to produce reasonable results when the learning rate is too large, due to the optimization process failing to converge.

\added{The test results in this section were obtained by applying the trained model to the original noisy data $\bmy$ rather than to the corrupted version $\bmyh$. Using $\bmyh$ slightly reduces the performance due to the influence of the auxiliary vector $\bmz$, as shown in Table~\ref{tab:yoryhat}, which compares the results under Poisson-Gaussian mixed noise. However, the performance gap is small (less than $0.1$ dB) due to the small variance of $\bmz$.}

\begin{table}[]
    \centering
    \addtolength{\tabcolsep}{-1.6pt}
    \caption{Model performance on $\bmy$ and $\bmyh$ at test time (mixed noise)}
    \label{tab:yoryhat}    
    \begin{tabular}{c|cc|cc|cc}
\toprule
\added{$(\lambda, \sigma)$} & \multicolumn{2}{c}{\added{$(15, 1.5)$}} & \multicolumn{2}{c}{\added{$(30, 3)$}} & \multicolumn{2}{c}{\added{$(60, 6)$}} \\
\added{Test data} & \added{BSD68} & \added{SET12} & \added{BSD68} & \added{SET12} & \added{BSD68} & \added{SET12}  \\ 
\midrule
\added{Original data $\bmy$} & \added{25.95} & \added{26.82} & \added{26.92} & \added{27.93} & \added{27.70} & \added{28.85} \\
\added{Corrupted data $\bmyh$} & \added{29.91} & \added{26.79} & \added{26.89} & \added{27.89} & \added{27.66} & \added{28.78} \\
\bottomrule
    \end{tabular}
\end{table}

\section{Conclusion.} 
We propose an approach for assessing the statistical noise consistency of existing denoising schemes and a learning-based refinement method to improve the results without using detailed information about the underlying noise distribution or any clean data. The method is flexible and achieves high-quality denoising results by refining the results of basic denoisers. The experiment results demonstrate its ability to correct errors in the denoising results and recover the lost information in the outcomes of existing denoising approaches, leading to small performance gaps compared to the supervised methods. However, the method requires training an additional network $G_\omega$ and hence extra computational cost. The extra cost can be outweighed by the added flexibility and capacity for improving existing results in situations where the standard pipelines can not be applied. Moreover, we focus on conditionally pixel-wise independent noise which is an important class of noise in applications. %

\noindent
\textbf{Acknowledgements}: Resources of ACRC at the University of Bristol and the Isambard-AI National AI Research Resource.

{\appendices
\section{Proof of Proposition \ref{prop:1}}
Proposition \ref{prop:1} states that the conditional expectation of $f(\bmz_i)$ can be expressed as follows: 
\begin{equation}\label{eq:fz2-S}
    \bbE_{\bmz_i} \qut{ f(\bmz_i) \mid  \bmyh } = \int p\qut{\bmx_i \mid \bmyh} G_\omega(\bmx_i, \bmyh_i) d \bmx_i,
\end{equation}
for some function $G_\omega$. As described in the paper, if the following equality holds, 
\begin{equation}\label{eq:E100}
\bbE_{\bmz_i}\qut{ f(\bmz_i) \mid \bmx_i, \bmyh }= \bbE_{\bmz_i}\qut{ f(\bmz_i) \mid \bmx_i, \bmyh_i },
\end{equation}
then one can set $G_\omega:=\bbE_{\bmz_i}\qut{ f(\bmz_i) \mid \bmx_i, \bmyh_i }$, and \eqref{eq:fz2-S} follows. 

In the remaining part of this section, we will prove the equality \eqref{eq:E100}. To simplify the notations, we will use $\bmz_i^c$ (resp. $\bmy_i^c$, $\bmyh_i^c$, and $\bmx_i^c$) to denote all elements of $\bmz$ (resp. $\bmy$, $\bmyh$, and $\bmx$) except the $i^{\rm th}$ element.

\textit{Step 1.}  
We start by defining the $\bmx_i$-dependent quantity $\xi$ as follows:
\begin{equation}\label{eq:E110}
\xi:=p\qut{\bmz, \bmyh \mid \bmx_i}.
\end{equation}
By applying the Bayesian Theorem, we can rewrite $\xi$ into the following form
\begin{equation}\label{eq:E111}
    \xi = p\qut{\bmz, \bmyh \mid \bmx_i} = p\qut{\bmz_i \mid \bmx_i, \bmz_i^c, \bmyh} p\qut{\bmz_i^c, \bmyh \mid \bmx_i}.
\end{equation}

\textit{Step 2.} Based on the fact that $\bmyh = \bmy + \bmz$, we have an equavilent expression for $\xi$:
\begin{equation}\label{eq:E101}
\begin{split}
\xi = p\qut{\bmz, \bmy \mid \bmx_i} & = p\qut{\bmy \mid \bmx_i} p\qut{\bmz \mid \bmx_i, \bmy}, \\
\end{split}
\end{equation}
where we use the Bayesian theorem in the second equality. The first term on the right hand side of \eqref{eq:E101} takes the form:
\begin{equation}\label{eq:E112}
p\qut{\bmy \mid \bmx_i} = p\qut{\bmy_i \mid \bmx_i} p\qut{\bmy_i^c \mid \bmx_i},    
\end{equation}
which is a consequence of the conditional independence in Assumption \ref{Assumption1}. The second term on the right hand side of \eqref{eq:E101} can be written as 
\begin{equation} \label{eq:E113}
\begin{split}
p\qut{\bmz \mid \bmx_i, \bmy} & = p\qut{\bmz_i \mid \bmx_i, \bmy} p\qut{\bmz_i^c \mid \bmx_i, \bmy} \\
&  = p\qut{\bmz_i \mid \bmx_i, \bmy_i} p\qut{\bmz_i^c \mid \bmx_i, \bmy_i^c}    
\end{split}    
\end{equation}
This is due to Assumption \ref{Assumption2}. Finally, combining the Equations \eqref{eq:E101}, \eqref{eq:E112}, and \eqref{eq:E113}, we obtain
\begin{equation}\label{eq:E121}
\begin{split}
    \xi & = \quts{p\qut{\bmy_i \mid \bmx_i} p\qut{\bmz_i \mid \bmx_i, \bmy_i}} \cdot \quts{ p\qut{\bmy_i^c \mid \bmx_i} p\qut{\bmz_i^c \mid \bmx_i, \bmy_i^c} }  \\
    & = p\qut{\bmz_i, \bmy_i\mid \bmx_i} p\qut{\bmz_i^c, \bmy_i^c\mid \bmx_i} \\
    & = p\qut{\bmz_i, \bmyh_i\mid \bmx_i} p\qut{\bmz_i^c, \bmyh_i^c\mid \bmx_i}.
\end{split}
\end{equation}
The last equality follows from the fact that $\bmyh_i = \bmz_i + \bmy_i$ and $\bmyh_i^c = \bmz_i^c + \bmy_i^c$. 
Equation \eqref{eq:E121} implies that $(\bmz_i, \bmyh_i)$ and $(\bmz_i^c, \bmyh_i^c)$ are independent conditioned on $\bmx_i$, hence $\bmyh_i$ and $(\bmz_i^c, \bmyh_i^c)$ are conditionally independent given $\bmx_i$. This implies that 
\begin{equation}\label{eq:E120}
p\qut{\bmz_i^c, \bmyh_i^c, \bmyh_i \mid \bmx_i} = p\qut{\bmyh_i \mid \bmx_i} p\qut{\bmz_i^c, \bmyh_i^c\mid \bmx_i}     
\end{equation}
Now, we continue simplifying the expression in Equation \eqref{eq:E121}:
\begin{equation}\label{eq:E122}
\begin{split}
    \xi & = p\qut{\bmz_i\mid \bmyh_i, \bmx_i} p\qut{\bmyh_i \mid \bmx_i} p\qut{\bmz_i^c, \bmyh_i^c\mid \bmx_i} \\
    & = p\qut{\bmz_i\mid \bmyh_i, \bmx_i} p\qut{\bmz_i^c, \bmyh_i^c, \bmyh_i \mid \bmx_i} \\
    & = p\qut{\bmz_i\mid \bmyh_i, \bmx_i} p\qut{\bmz_i^c, \bmyh \mid \bmx_i},
\end{split}    
\end{equation}
in which the second equality is based on \eqref{eq:E120}. 

\textit{Step 3.} Finally, \eqref{eq:E122} holds for all possible values of $\bmz_i^c$, and without loss of generality, we assume that $p\qut{\bmz_i^c \mid \bmx_i, \bmy_i^c} \neq 0$. We can conclude from Equations \eqref{eq:E111} and \eqref{eq:E122} that 
\[
p\qut{\bmz_i \mid \bmx_i, \bmyh, \bmz_i^c} = p\qut{\bmz_i\mid \bmx_i, \bmyh_i},
\]
and consequently $p\qut{\bmz_i \mid \bmx_i, \bmyh} = p\qut{\bmz_i\mid \bmx_i, \bmyh_i}$.
This gives the desired results in \eqref{eq:E100}, and the proof is completed. \hfill\ensuremath{\square}
}

\added{While $\bmn$ is assumed to be pixel-wise independent in Proposition \ref{prop:1}, an analogous result holds when it has a correlated component $\bmn^{\parallel}$. The next Proposition suggests that the same approach applies to the uncorrelated part of the noise.
\begin{Proposition}\label{prop:2}
    If $\bmn = \bmn^{\parallel} + \bmn^\bot$ where $\bmn^\bot$ is pixel-wise independent given $\bmx+\bmn^{\parallel}$, then under Assumption \ref{Assumption2}, there exists a two dimensional function $G_\omega: \bbR^2 \rightarrow \bbR$, such that
    \begin{equation}\label{eq:fz_c}
        \bbE \qut{ f(\bmz_i) \mid  \bmyh } = \int p\qut{\bmx_i + \bmn^{\parallel}\mid \bmyh} G_\omega(\bmx_i + \bmn^{\parallel}, \bmyh_i) d \bmx_i.
    \end{equation} 
\end{Proposition}
Hence our method can handle the uncorrelated component $\bmn^\bot$ in the same way it treats pixel-wise independent noise.
}

\bibliographystyle{plain}
\bibliography{references}

\vfill

\end{document}